# A Framework for Real-Time Face and Facial Feature Tracking using Optical Flow Pre-estimation and Template Tracking

Master thesis E.R. Gast

LIACS, Leiden University

April 2010

Supervisor: Michael S. Lew

# Abstract

This thesis presents a framework for tracking head movements and capturing the movements of the mouth and both the eyebrows in real-time. We present a head tracker which is a combination of a optical flow and a template based tracker. The estimation of the optical flow head tracker is used as starting point for the template tracker which fine-tunes the head estimation. This approach together with re-updating the optical flow points prevents the head tracker from drifting. This combination together with our switching scheme, makes our tracker very robust against fast movement and motion-blur. We also propose a way to reduce the influence of partial occlusion of the head. In both the optical flow and the template based tracker we identify and exclude occluded points. When the position and orientation of the head is known, we use this together with the 3D model to find the mouth and eyebrow movements. We use the head estimation to create a rectified image (RI) patch of the mouth and eyebrow regions. The mouth tracking is done by finding the right mouth model deformations so that the initial (closed) mouth is reconstructed. Eyebrow tracking is done by minimizing an error function which depends on template similarity and an eyebrow shape constraint. To evaluate our framework and its different trackers, we have conducted experiments using the well known Boston Head Tracking database [1] and our own dataset. The results show that our tracker can successfully track the head of different people. Also, the combination of the optical flow and the template based trackers reduces the number of head losses significantly when having fast movement or motion-blur. Excluding occluded points from the tracking process makes the tracker less sensitive to occlusion and makes the "head lost" detection more robust. The results of the mouth and eyebrow trackers show that basic expressions can be captured and that they can recover very well after erroneous tracking. Furthermore, we have evaluated the computation time of each component of the framework and measured an average total computation time of 35 ms, that is 29 fps.

# Table of Contents

# 1  Introduction

Head tracking has been a topic of research for a long time. Unfortunately, it is a very difficult problem to solve. This is because you have to deal with problems such as illumination changes, occlusions, pose changes, fast movement etc. In fact, tracking the head is probably more difficult than the more general object tracking problem because of its smooth surface, face deformations due to facial expressions, different facial hair and the possible glasses. Various methods have been proposed like using facial-features, Active Appearance Models and templates to track the face. Some of the methods can successfully track the head and some are also able to capture the movement of the facial features e.g. mouth and eyebrows. But they always require specific circumstances, training or are not capable of real-time performance. In this thesis we propose a framework to track head movements and capture mouth and eyebrow movements in real-time. Also the influence of illumination changes, occlusions, different faces and fast movement is minimized as much as possible.

To track the head and obtain its pose, we combine two different approaches. First we track the movement of projected model points using an optical flow algorithm. Using a minimization technique, model and tracked points are registered to acquire the head pose. Because this method alone suffers from drifting and occlusions an extra phase is introduced which tracks the head based on a template tracking approach. The extra phase can correct the first estimation and produces a better estimation. To improve robustness, a method to reduce drifting and a method to detect and remove occluded points in the optical flow and in the template tracking phase is proposed.

When the pose of the head is known we are able to track the mouth and the eyebrows. Because we know the head pose we also roughly know the location of the mouth and eyebrows. In order to track the mouth and eyebrows we create rectified image patches of the mouth and eyebrows regions. The rectified images make it possible to reduce search complexity and therefore the algorithms are more robust and less computational intensive. To track the mouth, a template tracking approach combined with a model-based approach is proposed. One of the benefits of this approach is that it can handle the mouth shape constraint very efficiently. Tracking the eyebrows is done by minimizing an error function. The error function depends on a normalized image similarity function and eyebrow shape constraints.

This thesis is organized as follows. Section 2 reviews the different methods and the current state-of-the-art in head and facial feature tracking. Section 3 gives some background information about techniques and methods we use in our framework. We give a full description of our framework in Section 4. In this section, the head, mouth, and eyebrows trackers are described. Section 5 shows the results of the different experiments. We discuss the obtained results in Section 6. Finally, in Section 7 we give a conclusion of our work.

## 2   Literature Study

A lot of work has been done in the field of head tracking. There are a lot of different approaches and methods proposed. All of these methods have their own advantages and disadvantages. There is no approach that can do perfect head tracking i.e. that it can handle different kind of occlusions, illumination changes, rotations and fast and slow movements and all in real-time. Most of the methods only address one or two of the important problems in face tracking. Also many of them need to be trained or initialized manually. In this chapter we discuss the most important methods. This chapter is divided into two parts. The first part is about estimating the pose of the head, the second part is about tracking facial features.

## 2.1   Head pose estimation

There are a lot of different approaches to estimate the head pose, therefore we only discuss the most important and promising ones; Pose estimation using facial features, Active Appearance Model, motion-based, and texture-based methods.

### 2.1.1   Head Pose Estimation using Facial features

Head pose estimation using facial features uses the location of certain facial features like the mouth, nose and eye to determine the head pose. This is very similar to the way humans 'estimate' the head pose of someone. One of the benefits of estimating using facial features is that it is a very intuitive and simple way of estimating the pose. An example of the simplicity is the approach of *Yingjie Pan et al.* [2]. They use seven points located on the mouth, nose and eyes to directly calculate the pose using the position changes of these points. Their approach can estimate the head pose but it is not very robust when dealing with for example occlusions and different lightning conditions. This is due to the fact that it completely depends on the accuracy of the detected features. If one feature is a false positive then the pose estimation will fail. With this approach it is very difficult to deal with different faces, because it uses the relationships between the features to estimate the pose. *Quang Ji* [3] uses only the pupil locations and an ellipsoid face tracker in combination with an IR (infra-red) camera to get their pose estimate. They first detect the pupils of the eyes, then they use the pupil locations as a starting point for their ellipsoid face tracker. After the ellipse is fitted onto the face they calculate the pose using the position of the pupils and the shape of the ellipse. They achieved good results. One of the reasons the algorithm achieved good results is because they used an IR camera, which let you detect pupils very easy and eliminates illumination problems. We won't use an IR camera, which makes the pose estimation using their approach still possible but won't perform as well.

*Wang & Sung* [4] proposed another method to estimate the pose. They make use of vanishing points. They estimate the pose by connecting the two outer and inner mouth corners and assuming that these lines are parallel. With this assumption you can obtain the head pose if the far-eye corners and the mouth corners are known. *Wang, Sung & Venkateswarlu* [5] proposed to use an expectation-maximization (EM) algorithm to deal with the variance of the facial model parameters i.e. the ratio between the eye-line and the mouth line. They found that their algorithm is reliable and is able to adapt to any individual. A downside of using vanishing points is that all the vanishing lines must be visible, this means that large rotations and occlusions are not possible. Also deformation of the face e.g. talking will have a big influence on the performance.

Pose estimation using facial features is a simple and fast method, and in specific cases it gives good results. The difficulty lies in detecting and tracking the features and handling missing or deformed features. Because of its limitations and inflexibility, head pose estimation using facial features is not a good approach for our problem.

## 2.1.2 Active Appearance Models (AAM)

A very interesting and popular approach is the use of Active Appearance Models (AAM). Active Appearance Models were first introduced by *Cootes & Edwards* [6]. They use statistical shape and texture models to form a combined appearance model. These shape and texture models are trained so they 'know' the modes of variation in shape and texture. The training set consists of face images which are manually annotated with landmark points to outline the face. Principal component analysis (PCA) [7] is applied to the vector of landmarks to reduce its dimensionality. Finally, the shape and texture modes are combined to form the AAM. Now the model has 'learned' the relationship between the displacement of the model parameters and the residual error. To fit the model, the current residuals are measured and the model is used to find a fit which reduces the residual error. The minimization of the residual error is done using a gradient-like approach and stops when the L2 norm of the error vector is below a certain threshold. An important thing to note is that the model must be initialized close to the observed face, otherwise it will converge to a (incorrect) local minimum. They got good results on tracking a face using 88 labeled training images. 19% failed to converge but the ones who converged where accurate. Unfortunately it did not run in real-time. This was because of the large number of iterations that the search algorithm needed to converge. Another problem was that all the landmarks had to be visible in each frame to track the face. This makes it not robust against occlusions. *Matthews & Baker* [8] introduced a fitting algorithm based on the inverse compositional image alignment algorithm [9]. Their algorithm achieved faster convergence, increased the number of convergences and reduced the computational costs of the algorithm. *Dornaika & Ahlberg* [10] improved the performance of the AAM algorithm by changing the search scheme. Instead of minimizing the distance between the image and its best approximation, they minimized the distance between the image and the synthesized image associated with the previous frame. This modified search scheme no longer depends on the number of face modes and therefore the performance improved.

A major problem of AAMs is that it is incapable of handling partial occlusion. *Gross, Matthews & Baker* [11] propose an algorithm to track faces that contain partial occlusions. They propose an algorithm to construct an AAM from training images with partial occluded faces and they propose the efficient robust normalization algorithm which is capable of handling occlusions. To deal with occlusions they introduce a robust error function. To make the algorithm cost efficient they assume that outliers are special coherent i.e. there are constant in each triangle. With this assumption a lot of computation can be moved outside of the search iteration loop. This gave the algorithm real-time performance. *Sung, Kanade & Kim* [12] proposed a method to make AAMs suitable for tracking faces across large pose variations and they also proposed a method to obtain good pose parameters for re-initialization when there is a false fit. In their algorithm they combine Active Appearance Models and a cylindrical head model. They use the pose parameters obtained from the cylindrical head as cue for the AAM fitting and re-initialization parameters. The cylindrical head tracking algorithm [13] they use performs better when tracking larger movements then the basic AAM. By combining the two trackers they improved the tracking rate and pose coverage significantly.

The big advantage of AAMs is that they can find a very precise location of the head and thus give a good pose estimation. But the main limitation is that its accuracy depends a lot on the training set that is used. This means that we need a lot of different faces and different poses in the training set if we want the AAM to work on different faces.

### 2.1.3 Motion-based Head Pose Estimation

A different but another intuitive method to estimate the head pose is tracking the head using the motion of the head or the motion of features on the face. A popular method to acquire the motion is to use an optical flow algorithm like described in [14]. In [15] they described a method for tracking a rigid head on video using optical flow. They used a technique called motion regularization together with an ellipsoid model as a base for tracking process. The main idea is to find the rigid motion of the head model that accounts best for the optical flow. First, the optical flow of each point is calculated and then they use a gradient descent technique to find the best motion of the head. The tracker in their test is very stable over a large number of frames and even while using sequences with a low frame rate and noisy images. *Decarlo & Metaxas* [16] proposed an optical flow based method to constraint the motions of a deformable model. They relaxed the constraint using a Kalman filter to deal with noisy data. To prevent drifting caused by the optical flow, they combined optical flow information and edge information. In [17] they tracked the projected model vertexes using optical flow. To estimate the pose they did a pose prediction at every iteration by solving the Least Squares of the motion parameters. The algorithm stops when a criterion is satisfied and the pose is found.

## 2.1.4   Template-based Head Pose Estimation

Template-based tracking is a well known tracking approach. The basic idea behind template tracking is to compare the observed image with a template image and find the position that gives the best match. One of the most important template tracking algorithms is the algorithm developed by *Lucas & Kanade* [18]. The goal of the Lucas-Kanade algorithm is to align a template image to an input image. They did this by minimizing the sum of squared differences (SSD) between the template and the input image that is warped back onto the coordinate frame of the template. The expression in equation (1) is minimized by the Lucas-Kanade algorithm. Where $W(x,p)$ is the warp of pixel $x$ using the parameters $\boldsymbol{p}$.

$$\sum_x [T(\boldsymbol{x}) - I\big(W(\boldsymbol{x},\boldsymbol{p})\big)]^2$$

(1)

The innovative thing they did was that they used the spatial intensity gradient of the image in combination with a minimization algorithm to find a good match. It was a lot faster than the previous algorithm because it examined far fewer  potential matches. It could also  be generalized to handle rotation, scaling and shearing. Because the original Lucas-Kanade algorithm had to re-compute the hessian matrix at every iteration, *Baker & Matthews* [9] proposed the Inverse Compositional Algorithm and reformulated the image alignment algorithm so the hessian could be pre-computed. This resulted in equation (2) .

$$\sum_x [T(W(\boldsymbol{x},\Delta\boldsymbol{p})) - I\big(W(\boldsymbol{x},\boldsymbol{p})\big)]^2$$

(2)

This new algorithm performs as well as the original Lucas-Kanade algorithm but it is less computational expensive. Template trackers like the Lucas-Kanade perform well, but unfortunately they are sensitive to illumination changes, occlusions and drifting. In spite of these shortcomings many template-based head trackers use the Lucas-Kanade algorithm as the base for their algorithm because of its excellence.

In [13] they created a 3D head tracker using a cylindrical model and a template tracker. They track the head by minimizing the SSD of the template and the image. The template is projected onto a 3D cylindrical model to better describe the head motion. They use an iteratively re-weighted least squares (IRLS) technique to deal with the non-rigid motion and some of the occlusions. To deal with self-occlusions and gradual illumination changes, they dynamically update and re-register the template. They achieved good results. *Chun, Kwon & Park* [19] also used a cylindrical model and template tracking to track the observed head but use a template update method to handle illumination changes. To produce a more realistic model they utilized a Gaussian radial basis function (RBF) to deform the 3D face model according to detected facial feature points from input images. This also gave their method the capability to do facial animations. *Matthews, Ishikawa & Baker* [20] and also used a template update method to handle illumination changes. Updating the template is a good way to handle illumination problems but there are some side effects. When using the template update method there is a big chance that template errors accumulate over many frames. When this happens, the tracker will start drifting. *Schreiber* [21] extends the template update algorithm [20] to deal with drifting. They use an algorithm similar to the algorithm of *Matthews et al.*[20], but instead of updating the template they use robust weights that are being updated from frame to frame. This proposed algorithm performs better than the original algorithm and it still remains fast. A side effect of the algorithm is that it will mask out strong edges which could be important features of the template. *Wolfgang* [22] evaluated two illumination-adaptive methods for template tracking; Brightness adaption by means of an illumination basis and a template update strategy. Their conclusion is that the template update method is more accurate and the preferred choice to use.

## 2.2   Facial feature tracking

Estimating the head pose and tracking its facial features (like the mouth and eyebrows) are closely related. To track facial features you often need to know the head pose so you have a rough estimate of the location of the features. As discussed in the previous section, it is also possible to estimate the head pose using the detected/tracked facial features. Because they are closely related, tracking facial features  can sometimes be included in the head tracking approach. For example, when an Active Appearance Model (AAM) is used, they often do not only track the head pose but also track facial features like the mouth and the eyebrows ([10], [23]). In the case of AAMs, including facial feature tracking (mouth and eyebrows) is often only a matter of adding additional (feature) parameters to the minimization process. Obviously, the facial features must also be annotated in the training set. [24] uses a more template based head tracker and is capable of simultaneously tracking the head and its facial features. Unfortunately, their approach only works in a controlled environment and did not achieve real time-performance .

*Tong, Wang, Zhu & Ju* [25], [26] present a multi-state hierarchical approach to track facial features in near frontal-view and half profile-view. A two-level hierarchy is proposed to characterize the global shape of the face and the local details of the facial components. To represent the feature points, Gabor wavelets and gray level profiles are combined. To deal with shape variations, they use multi-state local shape models. With these multi-state local shape models they can account for non-linear face deformations which a single-state model cannot account for. For example, they use three states (open, closed and tightly closed) for the mouth. To improve the shape constraints which are used for the feature search, they use an estimation of the head pose. To track the facial features they use a multi-modal tracking approach that dynamically estimates the feature component states and the position of the facial features. Their results show that their approach significantly improves the accuracy and robustness of the facial feature tracking under pose variations and face deformations.

*Zhu & Ji* [27] proposed a robust technique to detect and track facial feature points in real-time under various face orientations and deformations. Before tracking starts, they roughly locate the feature points in the image by positioning a face mesh on the observed face based on the located eye positions. Then, the point locations are refined by a nearest neighbor search approach. To track the points, first, Kalman filtering is used to give a prediction of the new position of every feature point. Next, a Gabor wavelet matching approach together with the predictions is used to detect the actual feature point position. To deal with face deformations, they dynamically update the Gabor wavelet coefficient at every frame. To refine the obtained results and handle feature occlusion they also included a shape-constrained correction method. Their results where good and they achieved real-time performance.

An often used approach to track features and especially the mouth, is too use active contours [28]. In [29] they use a combination of active contours and the Lucas-Kanade (LK) tracking algorithm to track the mouth. Initially, the mouth corners and vertical extrema of the lips are located using luminance, hue and gradient information of the area and contour. Then, the contours of the lips are traced and extracted by the active contour algorithm. The actual tracking is done with the LK-algorithm. The LK-algorithm tracks the initially detected points and is used as the initialization point for the active contour tracker in the next frame. This method works well, but only in a very controlled environment and is thus not very practical. In [30] they also extract the lip boundaries to track the mouth movements. They first extract the mouth region and then the outer and inner lip boundaries are traced and extracted. The outer lip boundary is extracted using a Gradient Vector Flow (GVF) snake in combination with parabolic templates as additional external force. The parabola templates are two parabolas describing the top and the bottom lip curve. The inner lip is tracked using a similarity function. Adding parabolic templates as external force improved the mouth tracking process. They achieved good results on the Bernstein sequence database. In [31] they initially detect the pupils, nose and mouth using a boosting algorithm and a set of Haar-features. When the features are located, the features are tracked using an optical flow base tracking algorithm. They also introduced a way to detect and recover tracking failure using model constraints and re-searching for features. The results where good compared to other algorithms and they state that is has strong potential as alternative method for building feature detection algorithms.

*Dahmane & Meunier* [32] present a modified phase-base approach to track facial features. For each point in the facial features, they generated a Gabor jet that independently corrects the node. In general it worked well, but it needs a large amount of manually annotated training samples. To overcome this shortcoming they extend [33] their approach by guiding the points using a main sub-graph of nodes. These nodes are not expected to be deformable and are thus a more reliable guide. This extension prevents the accumulation of tracking errors and thus prevents drifting.

*Su & Huang* [34] propose a facial feature tracking using a particle filter and believe propagation. Their contribution is that they extend the particle filter so that it can track multiple features simultaneously. Normally, an independent particle filter is used for every feature, but they introduce a spatio-temporal graphics model for tracking multiple facial features. Believe propagation is used to infer the spatial relationship between the different facial features. The relationships are learned beforehand using a large facial expression database. The results show that they can robustly track multiple facial features. Unfortunately, they have only tested their approach using a fixed camera.

## 2.3    Challenges and goals

One of the biggest challenges of this project is to create a complete framework that can do real-time and robust head and facial feature tracking in a way that it feels natural. This means that the tracking process must be precise enough to make it believable and so that it cannot be easily disrupted by external factors like occlusions. And if the tracker is disrupted it should be able to notice this and correct or re-start itself. Another challenge is to make the framework as independent as possible i.e. it relies as little as possible on for example training data or manual initialization, so that it works with different faces and is easy to use.

# 3 Background

## 3.1 Definitions

Throughout the whole thesis we use a number of definitions and a certain writing style. We denote bold capital letters as matrices, bold lowercase letters as vectors and regular lowercase letters as scalars. $\boldsymbol{M}$ is a transformation matrix and $W(\boldsymbol{p}, \boldsymbol{a})$ transforms points using the parameters specified in $\boldsymbol{a} = [\emptyset_x, \ \emptyset_y, \ \emptyset_z, \ s, t_x, t_y]$ .

$$\boldsymbol{M} = \ \boldsymbol{R}_x \boldsymbol{R}_y \boldsymbol{R}_z \boldsymbol{S} + \boldsymbol{T} \tag{3}$$

$$W(\boldsymbol{p}, \boldsymbol{a}) = \boldsymbol{M}\boldsymbol{p} \tag{4}$$

$\boldsymbol{M}$ is a $4 \times 4$ matrix, and $\boldsymbol{p} = \begin{bmatrix} x \\ y \\ z \\ 1 \end{bmatrix}$ is a column vector. $\boldsymbol{R}_x, \boldsymbol{R}_y, \boldsymbol{R}_z$ are the rotation matrices respectively around the x, y and z-axis. $\boldsymbol{S}$ is the scaling matrix and $\boldsymbol{T}$ is the translation matrix.

## 3.2 The CANDIDE model

In our framework we use the CANDIDE-3 model (Figure 1) to track the head and facial features. CANDIDE-3 is a parameterized face model specifically developed for model-based coding of human faces and is used in a wide range of articles ([10], [23], [35], [36]). It has a low number of polygons and consists of 113 vertexes. The model is controlled using local action units (AUs) which allows you to animate the face e.g. open the mouth or lift the eyebrows. With the Shape units (SUs) you can change the shape of the model e.g. the location of the mouth or eyes.

The shape of the model is given by a set of vertexes and triangles. You can represent the model as a 3N-vector $\bar{\boldsymbol{g}}$

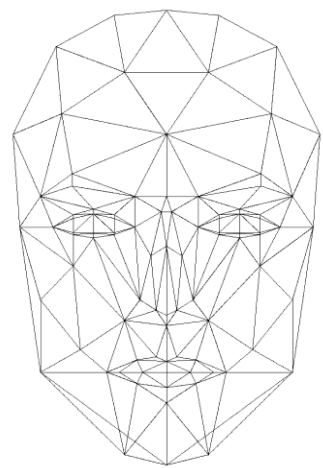

Figure 1: The CANDIDE-3 wireframe model.

containing the coordinates of the vertexes and where N is the number of vertexes. $\overline{\boldsymbol{g}}$ is the base shape of the model that can be reshaped to get $\boldsymbol{g}$ using:

$$\boldsymbol{g} = \overline{\boldsymbol{g}} + \boldsymbol{S}\sigma + \boldsymbol{A}\alpha \tag{5}$$

Where $\boldsymbol{S}$ and $\boldsymbol{A}$ are respectively the shape-units and the animation-units. $\sigma$ and $\alpha$ respectively contain the shape and animation parameters to reshape the base shape. To incorporate global motion the equation can be rewritten to:

$$\boldsymbol{g} = \boldsymbol{R}s(\overline{\boldsymbol{g}} + \boldsymbol{S}\sigma + \boldsymbol{A}\alpha) + \boldsymbol{t} \tag{6}$$

Where $\boldsymbol{R}$ is the rotation matrix, $s$ is the scale and $\boldsymbol{t}$ is the translation vector. Or it can be written as:

$$\boldsymbol{g} = \boldsymbol{M}(\overline{\boldsymbol{g}} + \boldsymbol{S}\sigma + \boldsymbol{A}\alpha) \tag{7}$$

Where $\boldsymbol{M}$ is an affine transformation matrix. The model can thus be parameterized by an affine transformation, shape and animation parameters. In this project we use one shape-unit to place the mouth plus an additional ten animation units, where four of them are used to animate the mouth and six to animate both eyebrows.

## 3.3   Levenberg-Marquardt method

The Levenberg-Marquardt (LM) algorithm [37], [38] is an iterative optimization algorithm which is particularly suited to find the minimum of a multivariate function that is expressed as the sum of squares. The LM algorithm can be thought of as a combination of steepest descent and the Gauss-Newton method. It combines the strengths of the two methods. Steepest descent is slow but it is guaranteed to converge and Gauss-Newton is fast but does not always converge. These properties make the LM method a popular optimization algorithm.

The LM algorithm gives a solution for the *nonlinear least squares minimization* problem i.e. minimizing a function in the form of:

$$E(\boldsymbol{a}) = \sum_{i=1}^{m} r_i^2(\boldsymbol{a}) \tag{8}$$

Where $\boldsymbol{a}$ is a vector of length $n$, $r$ is a function form $\mathbb{R}^n \to \mathbb{R}$ and $m \geq n$. Then $E$ can be rewritten as $E(\boldsymbol{a}) = \|r(\boldsymbol{a})\|^2$.

The LM algorithm chooses at each iteration an update $\delta\boldsymbol{a}$ to the current estimate $\boldsymbol{a}_k$ so that $\boldsymbol{a}_{k+1} = \boldsymbol{a}_k + \delta\boldsymbol{a}$ reduces the error function $E(\boldsymbol{a})$ i.e. $E(\boldsymbol{a}_{k+1}) < E(\boldsymbol{a}_k)$. The basis of the algorithm is a linear approximation to $E$ in the neighborhood of $\boldsymbol{a}$. For a small change in $\|\boldsymbol{a}\|$ you can write the following approximation using the Taylor expansion.

$$r(\boldsymbol{a} + \delta\boldsymbol{a}) \approx r(\boldsymbol{a}) + \boldsymbol{J}\delta\boldsymbol{a}$$

(9)

Where $\boldsymbol{J}$ is the $m \times n$ Jacobian matrix $\frac{\partial r(\boldsymbol{a})}{\partial \boldsymbol{a}}$ containing the first partial derivatives of $E$. The minimization is thus finding the update step $\delta\boldsymbol{a}$ which minimizes $r(\boldsymbol{a} + \delta\boldsymbol{a})$. This means that at each iteration we have to find the update $\delta\boldsymbol{a}$ which minimizes $\|r(\boldsymbol{a}) - \boldsymbol{J}\delta\boldsymbol{a}\|$. Differentiating the squares in (8) and equate with zero yields:

$$\boldsymbol{J}^T\boldsymbol{J}\delta\boldsymbol{a} = -\boldsymbol{J}^T r$$

(10)

When you solve this equation to $\delta\boldsymbol{a}$, than you get the *Gauss-Newton* update step:

$$\delta\boldsymbol{a} = -(\boldsymbol{J}^T\boldsymbol{J})^{-1}\boldsymbol{J}^T r$$

(11)

Levenberg and later Marquardt improved the algorithm by using a damped *Gauss-Newton method*. The damping factor $\mu$ which is adjusted at every iteration influences both the direction and the size of the steps. If the damping value is large, the step $\delta\boldsymbol{a}$ is near steepest descent direction which is good if the current iterate is far from the solution. If $\mu$ is small a *Gauss-Newton* like step is used which is good for the final stages of the iteration because of its quick convergence. In (12) you can see the damped equation.

$$(\boldsymbol{J}^T\boldsymbol{J} + \mu\boldsymbol{I})\delta\boldsymbol{a} = -\boldsymbol{J}^T r$$

(12)

$$\delta\boldsymbol{a} = -(\boldsymbol{J}^T\boldsymbol{J} + \mu\boldsymbol{I})^{-1}\boldsymbol{J}^T r$$

(13)

The LM algorithm stops when it meets one of the following conditions:

- If the right hand size of equation (12) i.e. $-\boldsymbol{J}^T r$ drops below a threshold $\epsilon_1$.

- If the relative change of the magnitude of $\delta\boldsymbol{a}$ drops below threshold $\epsilon_2$.

- Or if the maximum number of iterations is reached.

We based our implementation of the LM-algorithm on [39] and [40].

# 4 The Framework

In this section we will give a general outline and a detailed description of the methods we use. First we will give a short outline and then we will focus on head tracking and dealing with illumination changes and occlusions. After the head tracking we will describe how we track the mouth and eyebrows.

## 4.1 General outline

The framework can be divided into two parts; a head tracking and a facial features tracking part. The head tracking part is responsible for estimating the head pose i.e. position, orientation and scale of the head. The head tracking method is also divided into two phases, the first phase tracks the head using optical flow tracked points and the other phase tracks the head using a template tracking method. In the first phase, projected model points are tracked using an optical flow algorithm. These points are used to estimate a rough head pose. This is done by minimizing the distance between the tracked projected points and the corresponding model points. Minimizing the distance means recovering the model pose parameters. The advantage of this optical flow tracking approach is that it's capable of tracking the head especially when the subject is moving fast. The downside of the method is that it is very prone to drifting and occlusions. To account for this we also introduce a method to prevent drifting and to detect and remove outliers. To further improve the tracking we 'fine-tune' the pose with a template tracker. This template tracker compares the current image with a face template and tries to find the head pose by minimizing the distance between the current image and the face template.

We track the mouth and eyebrow movements using two different trackers. For both trackers, a rectified image of the regions is extracted. This rectified image 'reconstructs' the region of the feature as it is viewed from a frontal-view. The mouth tracker uses the rectified image to track four parts of the mouth; the two mouth corners, the upper and the lower lip. Instead of tracking these parts, the tracker tries to reconstruct the initial (closed) mouth. This approach eliminates most of the shape constraints which and also lead to a reduction of its complexity. The 'reconstruction' is done by adjusting the Animation Units (AUs) of the model.

The eyebrow tracker uses a similar approach as the head template tracker i.e. it tries to minimize a error function based on template similarity. But this error function does not only depends on template similarity but also on a shape constraint measurement. We minimize the error function to get three parameter values which describe the eyebrow movements. Each parameter value represents the displacements inside the rectified image of one of the three eyebrow parts. The three parts are the two eyebrow corners and the middle part of the eyebrow. Then, we translate the obtained parameters to AU parameters so we can animate the model.

## 4.2   Head tracking

The head tracking part is divided in two phases; an optical flow based tracking phase and a template tracking based phase. In the optical flow phase, points on the face are tracked using the Lucas-Kanade (LK) optical flow tracker [14],[41]. From the displacement of the tracked points it is possible to compute the movement of the head. For example, if you move the head to the left, all the points will have the same displacement and direction to the left, but if we rotate the head to the left (around the y-axis) some points will move faster than others. This can help with finding the head movement between two frames using the optical flow of the points. To find the new head pose we can treat it as a minimization problem where we want to estimate the pose of the model which accounts best for the tracked points. We do this by minimizing the distance between the points in the previous frame and the tracked point in the current frame. We can formulate this as an Iterative Closest Point problem [42], [43] where we already have the corresponding points pairs. Using this approach we can calculate the new pose of the head. Unfortunately, this approach alone is very prone to drifting and occlusions. This is mainly because we use optical flow to track the points. To handle this, we use an additional template based tracker. The template based tracker uses the pose estimate from the optical flow method as a starting point to do a more precise estimate. The template tracker 'fine-tunes' the head pose and also corrects most of the errors made by the optical flow tracker by re-projecting and updating the model points to prevent drifting. Unfortunately, the template tracker cannot correct all the estimation errors, so we also introduce a method to detect and remove outliers from the optical flow and the template tracker based method. Removing outliers makes the process more robust against illumination changes and occlusions.

The whole head tracking framework consists of some additional parts; initial model adaptation, template extraction and poor tracking detection. Before we start the tracking process we need to adapt the CANDIDE head model so it represents the face as much as possible. This is important for the extraction of the face template and for the accuracy of the feature trackers. To do the model adaptation and template extraction we detect the face using an object detector. We use the object detector which is initially proposed by Viola [44] and later improved by Lienhart [45]. We use the standard OpenCV implementation of this algorithm. For an outline of the head tracking framework see Figure 2.

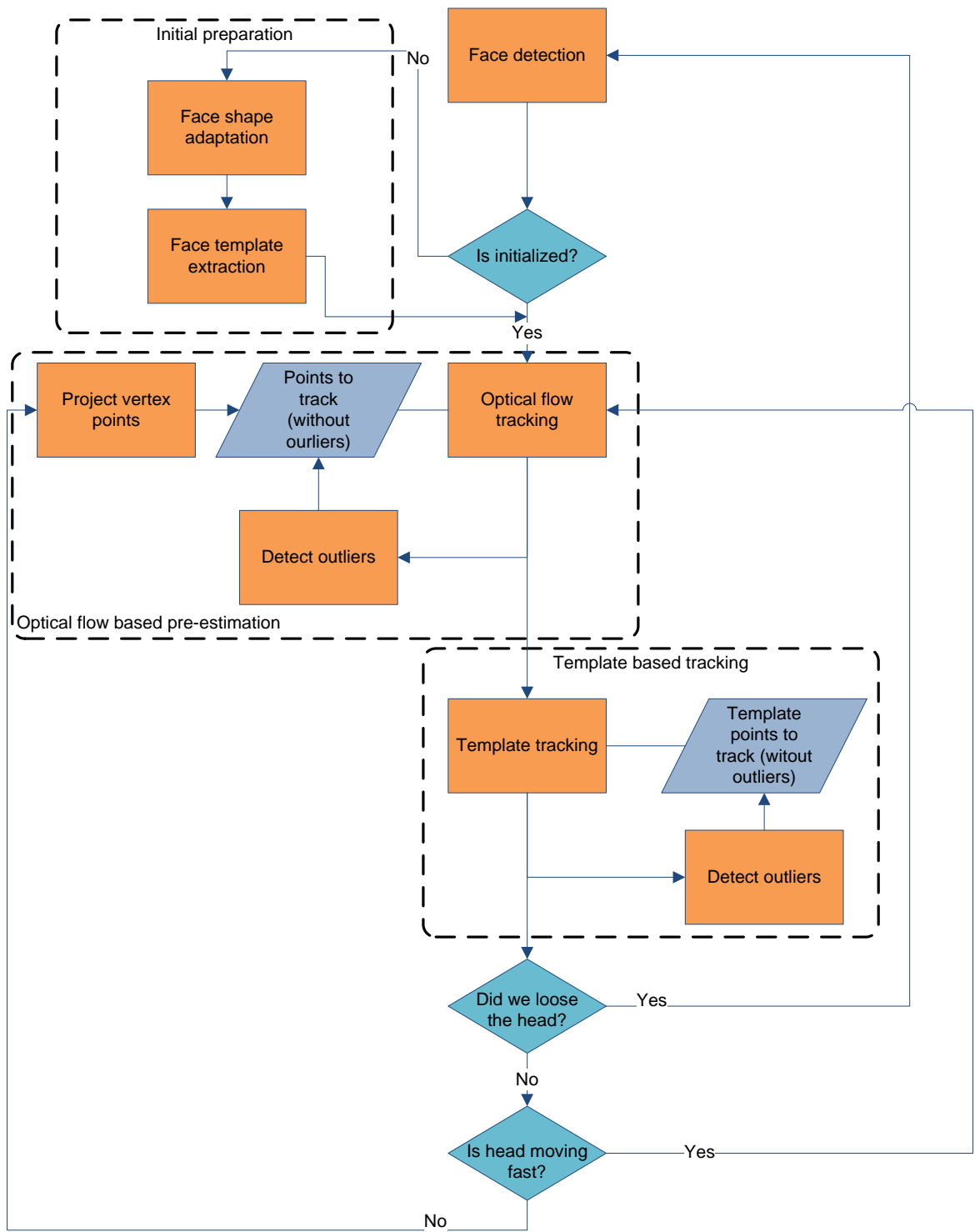

Figure 2: A general overview of the head tracking process.

## 4.2.1 Initial preparations

Before we can start tracking the face we have to know the initial position of the head and we have to make sure that the model resembles the face we want to track (the observed head) as much as possible. It is very important that the position of the models mouth is the same as that of the observed head. This is because we need a precise mouth location in order to make the mouth tracker work and it can also improve the head tracking performance significantly. Before we can determine the location of the mouth, we first need to know the position of the head and if it is rotated (around the z-axis). Locating the head is done using the object detector proposed by Viola [44] as mentioned before. To calculate the rotation around the z-axes we also detect the mouth and the two eye locations using Viola's algorithm. When we have the position of the eyes we can calculate the rotation (in radians) around the z-axis like this:

$$R_z = \cos^{-1}(\hat{e}_y) - \frac{\pi}{2}$$

(14)

Where $\hat{e} = \frac{eye_{right} - eye_{left}}{|eye_{right} - eye_{left}|}$ is the unit vector of the difference between the right and left eye center location. We also use the detected eyes to calculate the scale $s$ of the head model.

$$s = \frac{|\hat{e}|}{|d\boldsymbol{p}_{eyes}|}$$

(15)

Where $|d\boldsymbol{p}_{eyes}|$ is the distance between the models right and left eye centers. If the head is rotated then the detected center of the observed head is not the 'real' center. To calculate the 'real' center we again make use of the eyes. We calculate the correction by calculating the average distance between the eye locations of the model and the detected eye locations.

$$\boldsymbol{t}_{correction} = \frac{\left((s\boldsymbol{p}_{left} + \boldsymbol{t}) - \boldsymbol{t}_{leye}\right) - \left((s\boldsymbol{p}_{right} + \boldsymbol{t}) - \boldsymbol{t}_{reye}\right)}{2}$$

(16)

$$\boldsymbol{t}_{corrected} = \boldsymbol{t} - \boldsymbol{t}_{correction}$$

(17)

Where $\boldsymbol{p}_{left}$ is the center point of the models left eye and $\boldsymbol{p}_{right}$ is the center point of the models right eye, $\boldsymbol{t}$ is the detected head center vector i.e. head translation vector, $\boldsymbol{t}_{leye}$ is the detected left eye center and $\boldsymbol{t}_{reye}$ is the detected right eye center. When we have the 'real' head center we can calculate the position of the mouth i.e. how much we need to move the models mouth vertically. We use the candied shape units (SU) to move the mouth. How much the mouth has to move is calculated as follows:

$$p_{mout\ h} = f\,\frac{\boldsymbol{m}_y - (\boldsymbol{p}_y + \boldsymbol{t}_{corrected\ y})}{s}$$

(18)

Where $\boldsymbol{m}$ is the detected mouth center, $\boldsymbol{p}$ is the rotation and scale corrected mouth center of the model and $f$ is the mouth SU move factor i.e. $f$ defines the rate the SU moves. We use $f = 10$.

Now that we have adapted the model and we know the initial position, scale and rotation of the head model, we can start tracking.

### 4.2.2  Optical flow based pre-estimation

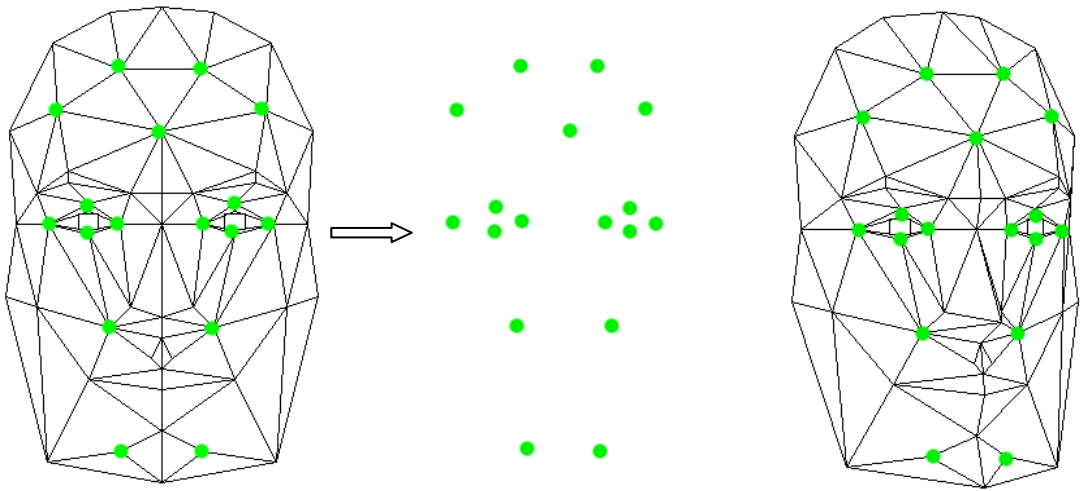

Figure 3: The general fitting approach using optical flow.

We estimate the head pose using the motion of certain points on the face. We use the motion of these points to find the model transformation that accounts best for the motion of these points. For this we use a 2D point registration approach (Figure 3). It is similar to the Iterative Closest Point (ICP) problem [43], but we minimize a slightly different error function then the one used in [43]. This gave us a better result. The points we track are projected vertex model points $\boldsymbol{p}_m$. This means that we track 113 points when using the CANDIDE-3 model (it has 113 vertexes). The projection of the points is given by:

$$\boldsymbol{p}_m = \boldsymbol{P}\boldsymbol{g}$$

(19)

Where $\boldsymbol{P}$ is the (orthogonal) projection matrix and $\boldsymbol{g}$ are the vertexes of the model. The points $\boldsymbol{p}$ will be tracked using the pyramidal implementation of the Lucas-Kanade optical flow algorithm [41]. We will denote the optical flow tracked points as $\boldsymbol{p}_{optf}$ and the projected model points as $\boldsymbol{p}_m$. Because we want to register the corresponding optical flow tracked points with their corresponding model points, we need a way to measure the alignment of the points. For this we use the absolute distance between the corresponding points. We minimize the total length between the points. So using the absolute distance between points we get the following residual and error function:

$$r(\boldsymbol{a}) = \sum_i^{N_p} \left| \boldsymbol{P}_{optf\,i} - W(\boldsymbol{p}_{m_i}, \boldsymbol{a}) \right| \tag{20}$$

$$E(\boldsymbol{a}) = r(\boldsymbol{a})^2 \tag{21}$$

Where $\boldsymbol{a} = [\emptyset_x,\ \emptyset_y,\ \emptyset_z,\ s, t_x, t_y]$ is the motion parameter vector with respectively the rotation angles, scale and translation parameters, $N_p$ is the number of points and $W(\boldsymbol{p}_{m_i}, \boldsymbol{a})$ (see equation (4)) transforms a point using the parameters in $\boldsymbol{a}$. $r$ is the residue and is a function from $\mathbb{R}^{N_p} \to \mathbb{R}$. We can estimate our parameters $\boldsymbol{a}$ by minimizing the error function (21). We found that minimizing the squared sum of absolute differences (20) is more robust for this particular problem than the sum of squared differences which is more often used. To minimize (21) we can make use of well known nonlinear optimization algorithms like Gradient-descent or Gauss-Newton, but we will use the Levenberg-Marquardt (LM) algorithm [37], [38]. The LM algorithm can be thought of as a combination of steepest descent and the Gauss-Newton method. It combines the strengths of the two methods. Steepest descent is slow but it is guaranteed to converge and Gauss-Newton is fast but is not guaranteed to converge. This combination makes the algorithm very fast and efficient and is therefore a popular optimization algorithm. You can find a more detailed description about the LM-algorithm in Section 3.3.

To minimize equation $E$, we need to compute the first order derivatives of equation $E$. We compute the derivatives of $E$ using forward finite differences [46]. An approximation of the first order derivative can then be computed with (22) when $h$ is small.

$$\frac{\partial E(a)}{\partial a} = \frac{E(a+h) - E(a)}{h}$$

$$(22)$$

Now we know how to calculate the first order derivative we can calculate the Jacobian matrix $\frac{\partial E(\boldsymbol{a})}{\partial \boldsymbol{a}}$ needed by the LM algorithm. In our case, the Jacobian matrix is a $1 \times N_{\boldsymbol{p}}$ matrix, where $N_{\boldsymbol{p}}$ are the number of pose parameters.

Unfortunately the Levenberg-Marquardt method is not robust against outliers. This is because the algorithm assumes that all observations (point distances) are correct. It does not expect outliers. Because it is very likely that there will be erroneous tracked points (outliers) in the observation data due to e.g. occlusion or fast movement, we need a way of dealing with them. This is usually done by adding a robust estimator to the error function like the Lorentzian [46] or Huber [47] robust estimator. Adding a robust estimator to the error function ensures that the minimization is less influenced by the outliers. We choose not to use a robust estimator because we noticed that it did not perform as well as we would like in our case. One of the reasons for this is that when we add a robust estimation it takes more iterations to minimize the error function and this results in a significant performance decrease. This happens because variant data has not as much influence as before. This is beneficial if the variations are outliers, but when they are not, important variations are neglected which will result in a slower or erroneous convergence. Another reason is that when we add a robust estimator, the tracker does not function well when the points are moving fast (i.e. when there is fast head movement), this is because fast movement gives a more variant data set due to optical flow tracking errors which then results in 'neglecting' important variations. Because handling fast head movements was the main reason to include an optical flow based head tracker we did not use a robust estimator but instead we detect and remove the outliers prior to the minimization process. We detect the outlier using this criterion:

$$outlier(x) = \begin{cases} outlier, & |(x^2 - M)/\sigma| > c \\ inlier, & else \end{cases}$$

$$(23)$$

Where $x$ is the distance between a point pair, $M$ is the median of all the point distances, $\sigma$ is the standard deviation of the distances and $c$ is a threshold value, we use $c = 2.0$ in our implementation. We use the median instead of the mean of the distances, because the median is more robust to outliers. The median has a higher breakdown point [47] than the mean, which means that the median is not influenced as much by outliers as the mean is. The mean has a breakdown point of 0 which means that even one outlier can ruin the mean. The median has a breakdown point of 0.5 which means that $\lfloor(n-1)/2\rfloor$ ($n$ are the number of observations and $\lfloor\cdot\rfloor$ denotes the floor operator) of the observation can be an outlier without making the median arbitrarily bad. This makes the median a better choice to use.

Unlike the robust estimator approach we don't detect and remove the outliers at every iteration of the minimization algorithm, but we detect and remove them prior to the minimization as shown in Figure 4. Also, we don't use the outliers of the current tracked points, but we use the outliers of the previous tracked points. The advantage of this is that we know for sure (assuming that the previous estimate is correct) which points are outliers and therefore there is no need to re-evaluate the points in every iteration like a robust estimator does. Using the outliers from the previous tracked points has the disadvantage that it is always outdated, this means that sudden changes (occlusions) won't be detected as outliers. Another problem with this outlier removal approach is that it is possible to mistake outliers for correctly tracked points and vice versa when there are a lot of erroneously tracked points. Luckily this does not happen very often.

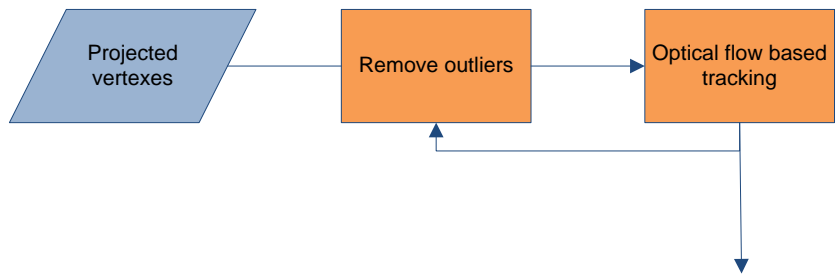

Figure 4: Outliers in the optical flow based tracker.

### 4.2.3 Template based pose estimation

The template based tracker is a more accurate face tracker and is used to correct and fine-tune the pose parameters obtained by the optical flow based approach. This tracker tracks a face template and recovers the corresponding pose parameters. The tracking approach is similar to other template trackers like [18] only we use a head model (CANDIDE), a normalized SSD and we remove the pixel outliers. Tracking the template can be seen as minimizing an error function like the sum of squared differences (SSD) or the sum of absolute differences (SAD) between the template and the observed image. We will introduce our error function after explaining the simpler SSD case. The basic error function $r$ we use is (24) , where $\boldsymbol{a}=[\emptyset_x,\emptyset_y,\emptyset_z,s,t_x,t_y]$ is the parameter vector, $\boldsymbol{x}$ is the $N_p$-vector of points and the residues $\boldsymbol{e}\in\mathbb{R}^{N_p}$.

$$\boldsymbol{e}=r(\boldsymbol{x},\boldsymbol{a})=I_T(\boldsymbol{x})-I\big(W(\boldsymbol{x},\boldsymbol{a})\big)$$

$$(24)$$

The 3D point $\boldsymbol{x}_i$ in $\boldsymbol{x}$ are points located on the head model and are the templates 'pixel' points we want to track. $W(\boldsymbol{x}, \boldsymbol{a})$ warps the points using the parameters in $\boldsymbol{a}$. $I_T(\boldsymbol{x})$ retrieves the pixel intensities of the template at the locations of the projected 3D points $\boldsymbol{x}$. $I(\boldsymbol{x})$ does the same only for the observed image. Using error function $r$ we can define the following minimization function to minimize the SSD of the template and the observed image.

$$\min_{\boldsymbol{a}} \|r(\boldsymbol{x}, \boldsymbol{a})\|^2$$

(25)

Now we can minimize (25) with a minimization algorithm, we use the Levenberg-Marquardt (LM) algorithm. Minimizing (25) with the LM-algorithm results in a head template tracker that converges in about 10 iterations to its minimum. Note that this only happens when an initial position close to the actual head position is given. Minimizing the squared differences works really well and is capable of tracking the head. Unfortunately, using only the squared differences makes the tracker very prone to illumination changes. Two often used approaches to handle illumination changes are image normalization and template updating [20], [21]. When using template updating it can be tricky to determine when to update the template. If we don't do this right it will result in a serious error accumulation as described in Section 2.1.4. Therefore we use the image normalization approach. We use a modified version of the *zero mean normalized sum of squared differences* (ZSSD-N) [48] to include normalization. While experimenting we found that ZSSD-N performed better than the *normalized sum of squared differences* (SSD-N) or any other normalized similarity measure. ZSSD-N looks like this:

$$\text{ZSSD} - \text{N} = \frac{\sum_{(x,y)\in \boldsymbol{U}}[(I_T(x,y) - \overline{I_T}) - (I(x,y) - \bar{I})]^2}{\sqrt{\sum_{(x,y)\in \boldsymbol{U}}(I_T(x,y) - \overline{I_T})^2 \cdot \sum_{(x,y)\in \boldsymbol{U}}(I(x,y) - \bar{I})^2}}$$

(26)

$I_T(x, y)$ is the pixel intensity of the template image at pixel coordinates $x$ and $y$, $I(x, y)$ is the pixel intensity of the observed image and $\overline{I_T}$ and $\bar{I}$ denote respectively the average intensity of the template and the observed image. If we rewrite (26) to a normalized residual function $r_{norm}$ we get:

$$v = \sqrt{\sum_{i\in \boldsymbol{x}}(I_T(i) - \overline{I_T})^2 \cdot \sum_{i\in \boldsymbol{x}}(I(W(i, \boldsymbol{a})) - \overline{I(W(i, \boldsymbol{a}))})^2}$$

(27)

$$e_{norm} = r_{norm}(\boldsymbol{x}, \boldsymbol{a}) = \frac{[(I_T(\boldsymbol{x}) - \overline{I_T}) - (I(W(\boldsymbol{x}, \boldsymbol{a})) - \overline{I})]}{\sqrt{v}} \tag{28}$$

Then the minimization function is:

$$\min_{\boldsymbol{a}} \|r_{norm}(\boldsymbol{x}, \boldsymbol{a})\|^2 \tag{29}$$

Unfortunately, $r_{norm}$ is not very efficient. Calculating the average intensity of the template, the observed image and calculating $v$ in every iteration is very expensive. When using the LM algorithm, the residual function $r_{norm}$ is called $i * N_{\boldsymbol{a}}$ times (in worst case). Where $i$ is the number of iterations and $N_{\boldsymbol{a}}$ are the number of parameters to estimate, which is $N_{\boldsymbol{a}} = 6$ in our case. And with $r_{norm}$ takes four times longer than $r$ to compute and with both having a complexity of $\mathcal{O}(n^2)$. Changing the residual function $r$ to $r_{norm}$ will have a noticeable impact on the real-time performance. So we propose not to use the mean $\overline{I}$ and $v$ of the current image, but to use the mean $\overline{I}$ and $v$ of the previously fitted image, $\overline{I}_{t-1}$ and $v_{t-1}$ respectively. When we do this we will have to compute the mean and $v$ just once per minimization. Another advantage is that you calculate the normalization factor using a correct face image. But because we use the image information of the previous image, it cannot handle sudden (per frame) illumination changes. This is actually not really a problem because this occurs only in exceptional situations.

$$e_{norm} = r_{norm}(\boldsymbol{x}, \boldsymbol{a}) = \frac{[(I_T(\boldsymbol{x}) - \overline{I_T}) - (I(W(\boldsymbol{x}, \boldsymbol{a})) - \overline{I_{t-1}})]}{\sqrt{v_{t-1}}} \tag{30}$$

Now we propose a way of detecting and eliminating outliers, for example caused by occlusions. When outliers are detected they will be excluded from the minimization process. If we exclude points (outliers) from the minimization, it means that $v_{t-1}$ which is used to calculate the residual function is not correct anymore. This is because $v_{t-1}$ is calculated when the detected outliers where not yet excluded. To correct for this we need to alter (27) into (31).

$$v = \sqrt{c \sum_{i \in \boldsymbol{x}} (I_T(i) - \overline{I_T})^2 \cdot \sum_{i \in \boldsymbol{x}} (I(W(i, \boldsymbol{a})) - \overline{I(W(i, \boldsymbol{a}))})^2} \tag{31}$$

$$c = \frac{N_{p_{t-1}} N_{p_T}}{N_p{}^2} \tag{32}$$

Where $N_p$ are the number of points to track in the current minimization i.e. the template points without the outlier points. $N_{p_{t-1}}$ are the number of points used in the previous minimization and $N_{p_T}$ are the total number of points in the template. If we use (31) instead of (27), we can change the amount of points we track without completely recalculating $v_{t-1}$ and thus efficiently exclude the outliers. To detect the outliers we use the following criteria:

$$outlier(x_i) = \begin{cases} outlier, & x_i > \sigma_x \\ inlier, & else \end{cases} \tag{33}$$

$$x = e_{norm}^2 \tag{34}$$

$$\sigma_x = \sqrt{\frac{\sum_{i \in x}(x_i - median_x)^2}{N_x}} \tag{35}$$

When the squared residue of a point $x_i$ is more than the standard deviation $\sigma_x$ of $x$, we say that the point is an occluded or badly illuminated point and is thus an outlier. The points that are detected as outliers are excluded in the next template tracking. After every successfully tracked head we re-determine the occluded points from $x$ and exclude them from the next tracking step. The benefit of excluding the occluded points from the previous frame is that you are almost certain that the points are really occluded and are not a variation in the face. The drawback, however, is that it is possible that occluded points are not yet seen as outliers and thus are used to track the template. Luckily this does not happen very often and it is thus not really a problem.

Another important aspect of the tracker is the selection of the template points. If we use too many points it will increase the computation time significantly and if we use too few points the tracker will be very fast but not very accurate. Also, we want to keep the computation time of the tracker constant at different template scales to make it scalable when using different image resolutions. There are a lot of ways how to choose the template points. You can for example use a feature detector like the Harris corner detector [49], the SIFT algorithm [50] to locate good points, random points or project a grid of points. We don't use any feature detector but use a simple grid of points which we project onto the model to find their 3D location on the model. The advantage of projecting a grid of points is that you can easily adjust the grid size and density and thus easily change the resolution of the template. When using a feature detector you don't have this kind of control. This makes the grid of points better scalable than the other approaches and our preferred choice. Also, there are not many noticeable differences in tracking performance when using a feature detector or a point grid approach to select the template points. In Figure 5 you can see how the grid projection is done.

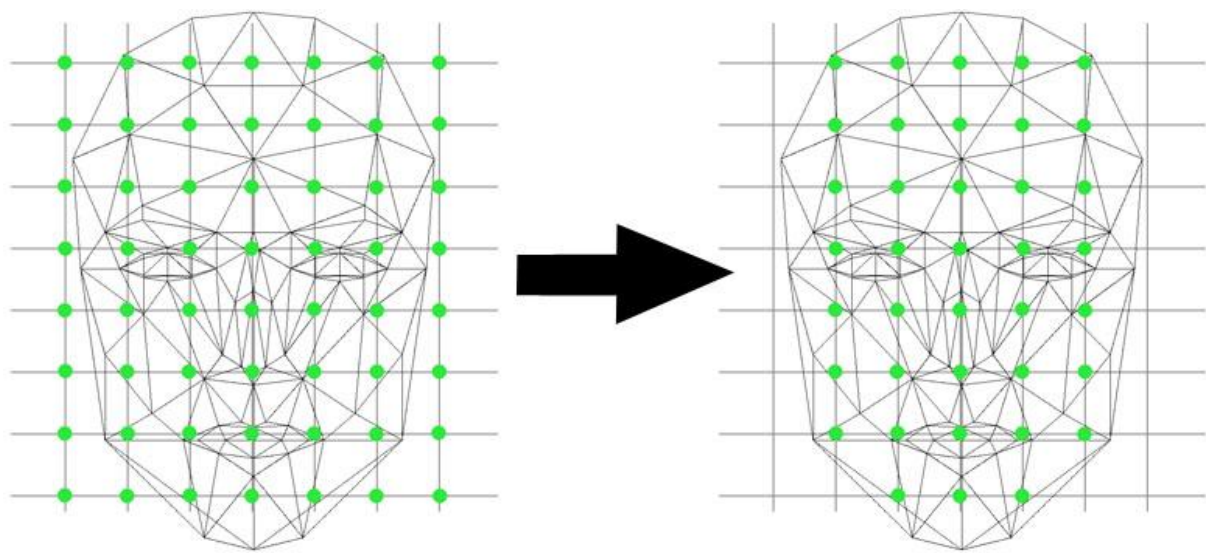

Figure 5: Projecting the grid of points onto the face model.

### 4.2.4   Combined approach

The strength of our approach is the combination of optical flow based tracking and a template based tracker to track the head. Each of the trackers have their own strengths and weaknesses. Optical flow base tracking is very good at tracking large and fast movements but it is not always very accurate. The template tracker is just the opposite. The template tracker is very accurate but the initial position must be close to the actual head location which makes it not very good in handling large and fast movements. So, to obtain the strengths of both trackers we combine them. First we track the head using the optical flow based trackers which gives us a rough estimate of the location and orientation of the head. This rough estimate will be close to the real head location but we can't assume that it is the actual head location. This is actually a perfect situation for the template base tracker, so after the rough estimate of the optical flow based tracker we fine-tune the estimate using the template based tracker. To prevent the optical flow based method from drifting and to correct its estimation error we re-project and update the optical flow points using the newly fine-tuned estimation of the head. Using this approach will improve the tracking and especially when there is moderate or slow head movement. It also improves tracking large and fast head movements, but unfortunately there is still a big chance to lose the head. This is due to re-projecting and updating the optical flow points. So to solve this problem we try to detect when it is likely that the tracker is not precise (i.e. when the head moves too fast). When this happens we make sure that the optical flow points are not updated for a couple of frames. So we update the optical flow points only when these conditions are met:

- $d_{optflow}$ is smaller or the same as a distance $d$ and the accuracy of the template tracker is acceptable i.e. when error (31) of the template tracker is below a certain threshold. We use $d = 10$.
- The standard deviation $\sigma_{optflow}$ is higher than a constant $c$ and the accuracy of the template tracker is acceptable. We use $c = 100$.

$$d_{optflow} = \left\| f(\boldsymbol{p}, \boldsymbol{a}_{optflow}) - f(\boldsymbol{p}, \boldsymbol{a}_{prev}) \right\| \tag{36}$$

Where $\boldsymbol{a}_{optflow}$ is the parameter vector obtained by the optical flow based tracker, $\boldsymbol{a}_{prev}$ is the parameter vector of the previous estimate of the tracker and $\boldsymbol{p}$ is an arbitrary point on the head model. $d_{optflow}$ gives the distance that the head has moved from the previous frame to the current frame solely based on the estimate from the optical flow tracker. So, we actually say that we cannot 'trust' the template head trackers estimate when the optical flow based tracker estimates a displacement of more than $d$. There is always a chance that the tracker will still lose the head due to for example too fast movement or occlusion. If this happens or when $d_{optflow} < d$ and the accuracy of the template tracker is not acceptable, we say that the tracker has lost the head. When the tracker has lost the head we re-locate the head using Viola's [44], [45] algorithm and reset the optical flow and the template based tracker and re-start both trackers.

## 4.3   Mouth and eyebrow tracking

To track the mouth and the two eyebrows we propose two methods; one to track the mouth and one to track each eyebrow. They will both make use of the head pose estimate obtained by the head tracker (Section 4.2). We use the head pose estimation to reduce the complexity of both the mouth and the eyebrow trackers. With the head pose information we create a rectified image (RI). To obtain the RI, head points inside a certain area are warped back to a frontal view (they are rectified). We also warp the image to a constant image size which makes template matching possible and this also makes the computation time constant (Section 5.3). Because we only warp head points which are located in a predefined the area, we simultaneously eliminate background pixels. Both the mouth and eyebrows are tracked inside the RI. When the tracking is done, the results are translated back to head coordinates.

### 4.3.1  Rectified image area

To calculate the rectified image $I_{rect}$ we first need to locate the points on the head model. We do this by projecting a $w \times h$ 2D point grid $\boldsymbol{H}$ onto the model (in frontal view), where $w \times h$ is the resolution of $I_{rect}$. The result will be a $w \times h$ 3D point grid $\boldsymbol{g_H}$ of points located on the head model.

$$\boldsymbol{H}_{x,y} = \begin{bmatrix} s_x x \\ s_y y \end{bmatrix} + \begin{bmatrix} t_x \\ t_y \end{bmatrix} \tag{37}$$

$$s_x = a_w / w$$
$$s_y = a_h / h \tag{38}$$

Where $x \in \{1, \dots, w\}$ and $y \in \{1, \dots h\}$, $a_w$ and $a_h$ are respectively the screen width and the height of the area we project.

$$\boldsymbol{g_H}_{x,y} = intersect(M, \boldsymbol{H}_{x,y}) \tag{39}$$

$intersect(M, \boldsymbol{p}_{x,y})$ retrieves the point of intersection of the projected 2D-point $\boldsymbol{p}_{x,y}$ using the head model $M$.

$$intersect(M, \boldsymbol{H}_{x,y}) = \begin{cases} \begin{vmatrix} \boldsymbol{H}_x \\ \boldsymbol{H}_y \\ 1 \end{vmatrix}, & \textit{if the projection of } \boldsymbol{H}_{x,y} \textit{ intersects } M \\ nothing, & otherwise \end{cases} \tag{40}$$

To obtain $I_{rect}$ we have to retrieve the interpolated intensities of the points in $\boldsymbol{g_H}$ with model parameters $\boldsymbol{a}$.

$$I_{rect\ x,y} = \text{Interpolate}(I, W(\boldsymbol{g_H}_{x,y}, \boldsymbol{a})) \tag{41}$$

Interpolate$(I, \mathbf{x})$ gives the interpolated value in image I with the coordinates $\mathbf{x}$.

### 4.3.2 Mouth tracking

Mouth tracking is one of the most difficult facial feature to track, since it has a complex shape, is highly deformable and can move fast. Here we propose a method to track the most important movements of the mouth i.e. opening/closing the yaw, raising/lowering the upper lip, moving the mouth corners horizontally and vertically. These mouth movements are modeled using four CANDIDE animation units (AUs). The tracker is based on a template tracking approach where we track both mouth corners, the upper and the under lip/yaw. A big problem of using template tracking for tracking the mouth, is maintaining a valid shape. Template tracking alone is not robust enough to preserve a valid the shape of the mouth. We can solve this by inverting the problem, we don't track the mouth parts, but we try to reconstruct the initial (closed) mouth template. Going from tracking the templates to reconstructing the initial template drastically reduces the search complexity and simplifies the shape constraint. Because the rectified image has the property that all points in the rectified image have a location on the head model, this means that the RI points can easily be translated to model points coordinates and vice versa. We use this property to 'reconstruct' the mouth. Since we know where the center of the mouth lies in the image and we know the shape of the mouth, we also know where the mouth corners and the upper and under lip lie in the image. And because we are reconstructing the mouth to its 'original' shape, we know that the tracked mouth parts have to end up at the same location in the reconstructed image as in the initial image. To do this, for every template, we locate the best match in a certain area around its initial location (in the original shape). When we find the location of the best match, we transform the model using the associated AUs so that this mouth part is transformed back to its original location i.e. reconstructing the mouth like shown in Figure 6.

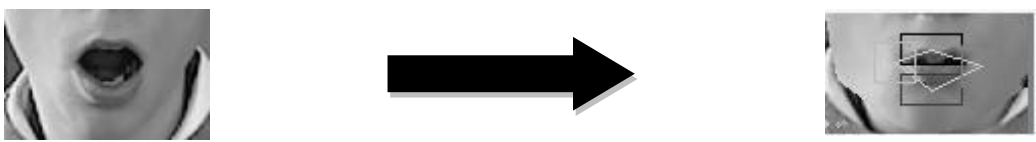

Figure 6: On the left the non-rectified mouth image, on the right the reconstructed mouth image.

Before we can track the mouth we first have to extract the templates of the mouth parts. We extract an upper lip, bottom lip, left and right mouth corner template (Figure 7). These templates are extracted in the preparation stage when the face is in frontal view. A rectified image of the mouth region is obtained and the templates are extracted from this image. To acquire the correct location of the templates, we project the mouth points from the model onto the image. It is important that the mouth corners are located correctly, so we project the mouth corners onto the rectified image and search for a more precise location using the Harris corner detector [49]. We only use the corner detector in a small region around the projected mouth corner, then we use the 'strongest' Harris corner as mouth corner location. After locating and extracting the templates, we have the four locations of the mouth part in the rectified image, we refer these points as $\boldsymbol{o}$.

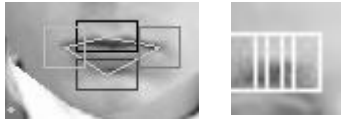

Figure 7: The template regions of the mouth (left) and the eyebrow (right) that are tracked.

For the actual tracking process, at every frame and for each template, a similarity map is calculated. For the upper and lower lip, this is a 1D similarity map in vertical direction. For the mouth corners, this is a 2D map in vertical and horizontal direction. The maps are filled with ZSSD-N values between the area location in the rectified image and the mouth template. This ZSSD-N is slightly different than the one used in Section 4.2.3. Here we don't use the intensities of the previous region for normalization, but we use the intensities of the whole template search region for normalization.

$$\text{Similarity} = \frac{\sum_{(x,y)\in T}\left[\left(I_T(x,y) - \overline{I_T}\right) - \left(I\left(d_x + i + x, d_y + j + y\right) - \overline{I}\right)\right]^2}{\sqrt{c\sum_{(x,y)\in T}(I_T(x,y) - \overline{I_T})^2 \cdot \sum_{(x,y)\in A}(I\left(d_x + x, d_y + y\right) - \overline{I})^2}} \quad (42)$$

$$c = \frac{N_T^2}{N_A^2} \quad (43)$$

Where $d_x$ and $d_y$ are respectively the horizontal and vertical offsets in the rectified image to the similarity map region, $i$ and $j$ are respectively the horizontal and vertical offsets in the similarity map region to the template match region. $\boldsymbol{T}$ and $\boldsymbol{A}$ are a set of 2D point coordinates respectively in the template and the similarity map region and $N_T$ and $N_A$ are the number of point coordinates in respectively the template and the total similarity map.

When the similarity maps are calculated, we select the best matches by searching for the smallest value in each similarity map. The position of the best match $d\boldsymbol{o}$ is then used to calculate the displacements of the mouth parts. The displacements are used to reconstruct the mouth image to the initial mouth image. To reconstruct the mouth image we have to deform the model. We use the AUs of the CANDIDE model to deform the mouth. We calculate the AU parameters as follows:

$$\alpha_i = (d\boldsymbol{o}_j - \boldsymbol{o}_j)s_i\boldsymbol{r}$$

(44)

Where $\alpha_i$ is the AU parameter for AU $i$, $d\boldsymbol{o}_j$ is the location of the best match of mouth part $j$ in the rectified image, $\boldsymbol{o}_j$ is the original location of mouth part $j$ in the rectified image, $\boldsymbol{r}$ is the rectified image pixel to model point ratio and $s_i$ is the rate AU $i$ deforms. Deforming the mouth parts is the same as 'moving' the templates back to their original position and thus reconstructing the mouth.

While template matching in combination with a rectified image performs good, it is still a problem when the head is rotated to much around its y-axis. If the head is rotated too much, one of the mouth corners will be partially or fully occluded. Since we use a simple similarity function, the template tracking will fail when this occurs. We solve this rotation problem in a very straightforward way. If the head rotation around the y-axis is not between certain thresholds, than the displacement of the occluded corner is the same as the non-occluded corner (the other mouth corner). We use a maximum rotation of $20°$ and $-20°$.

Algorithm 1: The mouth tracking algorithm.

**Begin**

1. Acquire the rectified image region, by projecting an area onto the model.
2. Set the initial locations of the mouth parts.
3. Extract the 4 mouth part templates $\boldsymbol{T}$.
4. **for**(every frame **F**)
   a) Get the rectified image $I_{rect}$ using the pose parameters $\boldsymbol{a}$.
   b) **for each**(template in $\boldsymbol{T}$)
      i) Calculate the similarity map of $\boldsymbol{T}$.
      ii) Find best match in the similarity map.
      iii) Calculate template displacement.
   c) Correct the locations of the mouth parts when the head rotation around the y-axis is between threshold $-r$ and $r$.
   d) Calculate the new mouth AU parameters, by adding the obtained displacements to the previous AU parameters. If the AUs exceed its minimum or maximum value, then it is set to respectively its minimum or maximum.
   e) Deform the head model using the mouth AUs.

**End**

We calculate the final mouth AUs by adding the acquired AU displacements of the mouth parts to their previous mouth AU parameter value. The final mouth AU parameter values are then checked if they are plausible. We do this by checking if they fall between a certain minimum and maximum value. If it exceeds the minimum or maximum, than the AU parameter value is set to the value it exceeded (minimum or maximum). The final outline of the mouth tracking algorithm is shown above.

### 4.3.3 Eyebrow tracking

Here we propose an eyebrow tracking approach. It is different than the mouth tracker and somewhat similar to the template head tracking approach. The main difference is that here we use an additional internal constraint to enforce the eyebrow shape. The basic idea is to track an eyebrow that consists of three parts; a middle part and two outer parts (eyebrow corners). Each part can move in vertical direction. This gives the eyebrow three degrees of freedom (3 DOFs). We animate the eyebrow movements by using an AU for each eyebrow part.

Like the mouth tracker, we use a rectified image to simplify the tracking process. Here we also track the different parts inside the rectified image. The results are then transformed back to model coordinates and the corresponding AU parameter values are calculated. For initialization, the three eyebrow parts are located on the rectified image and the associated template for each part is extracted. This template extracting approach is exactly the same as in the mouth tracker, so for more information see Section 4.3.2.

The extracted eyebrow template parts are tracked by minimizing an error function. This error function depends on template similarity and an eyebrow shape constraint. We define the error function as follows:

$$\boldsymbol{e}_i = E(i, \boldsymbol{a}) = \sqrt{w_i} \, |I(\boldsymbol{a}_i) - I_i| \qquad\qquad i \in T \qquad\qquad (45)$$

$$w_i = (1 + c \, Int_i(\boldsymbol{u})) \qquad\qquad (46)$$

$$Int_i(\boldsymbol{u}) = \left( \frac{\left\| \boldsymbol{u}_i - \cos^{-1}\left( \left( \frac{\boldsymbol{p}_{i-1} - \boldsymbol{p}_i}{|\boldsymbol{p}_{i-1} - \boldsymbol{p}_i|} \right) \cdot \left( \frac{\boldsymbol{p}_{i+1} - \boldsymbol{p}_i}{|\boldsymbol{p}_{i+1} - \boldsymbol{p}_i|} \right) \right) \right\|}{2} \right)^2 \qquad (47)$$

Where $w$ is a weight function, and $Int(\boldsymbol{u})$ is the internal energy that enforces the shape constraint. $\boldsymbol{u}$ is a vector with the optimal angles between two eyebrow parts, $\boldsymbol{a}$ is a vector with the displacement of the eyebrow part in the y-direction in the rectified image and the $|\cdot|$ operator denotes the dot-product between two vectors. We use a straight eyebrow model, so our shape angles are all 180°. If we calculate $Int(\boldsymbol{u})$ of one of the corner parts (which has only one neighbor part), we use a mirrored neighbor as its other neighbor to calculate the angle e.g. if we calculate the internal energy of the left eyebrow corner $i = 1$, we use $\boldsymbol{p}_{i-1} = \boldsymbol{p}_i + (\boldsymbol{p}_i + \boldsymbol{p}_{i+1})$ and with the right corner we use $\boldsymbol{p}_{i+1} = \boldsymbol{p}_i + (\boldsymbol{p}_i - \boldsymbol{p}_{i-1})$. $c$ in equation (46) is a constant to relax or increase the internal energy constraint. Increasing $c$ enforces a more rigid eyebrow. The $|I(\boldsymbol{a}_i) - I_i|$ part of equation (45) is the similarity measure between the rectified image and the template of eyebrow part $i$.

We track the eyebrow parts by minimizing equation (48) using the Levenberg-Marquardt (LM) algorithm (Section 3.3). The minimization result $\boldsymbol{a}$ is then transformed to model space and translated to AU parameter values. The conversion from rectified image coordinates to model coordinates and translating the results to AU parameters values is done in the same way as the mouth tracker in Section 4.3.2.

$$\min_{\boldsymbol{a}} \sum_{i=1}^{N} E^2(i, \boldsymbol{a}) \tag{48}$$

Where $N$ are the number of eyebrow parts (we use $N = 3$), $\boldsymbol{a}$ has a length of $N$ and contains the y-coordinates of the different eyebrow parts.

# 5 Experiments

To examine the accuracy of our proposed framework, we did a number of experiments. First we test the rotational accuracy of our tracker using a well known database. Second, we test the rotational and the vertical and horizontal movement accuracy. Third, we evaluate the performance of our tracker when it has to deal with a fast moving subject. Fourth, the robustness against occlusions together with the influence of outlier removal on dealing with occlusions is examined. Fifth, we visually inspect the performance of the mouth and eyebrow trackers. Finally, we measure the computation times of the different parts of the framework. In all experiments we use at least two different subjects to determine the influence of different faces on the performance.

In our experiments we try to give a quantitative evaluation of the performance of our tracker, but unfortunately this is not always possible. It requires an accurate ground truth which is difficult to acquire. There are a few methods to capture movements such as: manual annotation, magnetic sensors, motion caption and inertial sensors. All these capture methods are too complex or very expensive, except for the manual annotation method of course. So, for most of the experiments we use semi-manual annotated videos to examine the performance. The disadvantage of annotating the video image by hand is that it is not very accurate, humans are just not capable of determining the exact location and orientation. Because we developed this tracker not to get an exact estimate but to get a believable estimation of the position and orientation of the head, this is not a problem. Instead of a precise error measurement, it is more important to know if the tracker follows the correct general motion. To relate our performance to other research, we also compare our tracking results with the ground truth of the Boston University Head Tracking database [1].

Table 1 shows which tracker configuration we used for our experiments. The used hardware configuration is shown in Table 2.

Table 1: The tracker properties used.

| Property | Value |
|---|---|
| Video size | *640x480* |
| Number of optical flow tracked points | *113* |
| Average number of template points | *397* |
| Max. number of iteration for pre-estimation | *10* |
| Max. number of iteration for template tracker | *10* |
| Rectified image size used for mouth tracker | *100x60* |
| Rectified image size used for eyebrow tracker | *30x30* |

Table 2: PC configuration used to run the experiments.

| Hardware | |
| --- | --- |
| **CPU** | *Intel Core2 QUAD Q9550 2.84GHz* |
| **RAM** | *4.00 GB DDR2* |
| **OS** | *Windows 7 pro. 64-bit* |
| | *(program is compiled as 32-bit)* |
| **Webcam** | *Logitech Quickcam pro 9000* |
| **OpenCV** | *1.1pre1a (32-bit)* |

## 5.1  Head tracking

We performed three different experiments to evaluate the performance of our head tracker. First, we measure the rotational performance of our head tracker using the Boston University Head Tracking database [1]. Second, we measure the translational and rotational performance and its capability to handle fast movements. Finally, we test the robustness against occlusion.

### 5.1.1  Experiment 1: Rotation performance

We use 39 video sequences from the Boston database. The 39 videos are taking under uniform illumination and there are videos with five different subjects (see Figure 8). The subjects performed a wide range of different head rotations and translations. They used a "Flock of Birds" 3D magnetic tracker to capture the ground truth data. This magnetic tracker has an positional accuracy of 0.1 inches and an angular accuracy of 0.5° but only when there are no large metal or electronic objects in the room. Because the videos are captured in an environment with standard furniture and equipment, the accuracy will probably not be as precise as stated. In [51], from their own experience, they also argue that magnetic sensors are highly sensitive to noise and small amounts of metal in the environment. Despite the doubt of the precision of the captured ground truth, it is still a good data set to compare to, because we are mainly interested in the general motion and not a precise fit. Furthermore, the videos are captured with a resolution of 320x240 and with 30 frames per second.

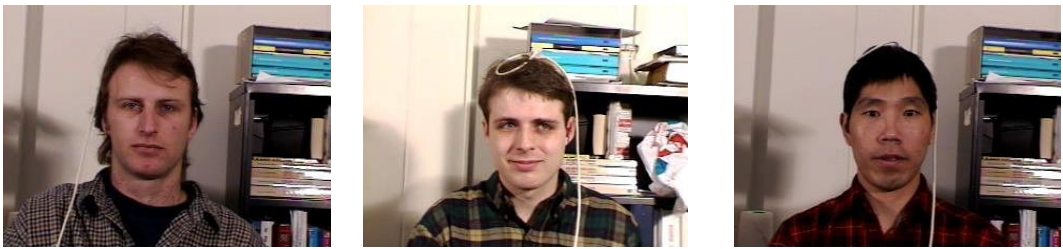

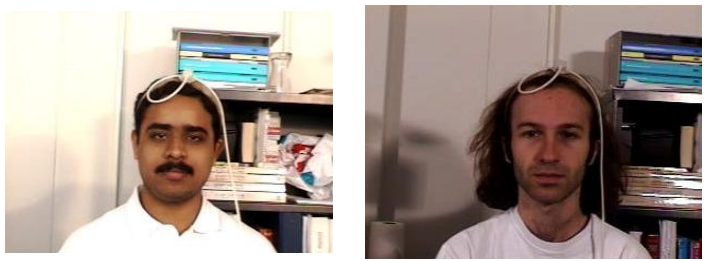

Figure 8: The five different faces used in de Boston University database.

This experiment is designed to evaluate the rotation accuracy of our tracker. The first thing the tracker does is detecting the face and adjusting the model, so that head model fits the head in the video. In most of the video sequences in the Boston dataset, our tracker does not detect the face in the first frame and does not have the same starting orientation as the ground truth. This means that we have to adjust our starting orientation to that of the ground truth.

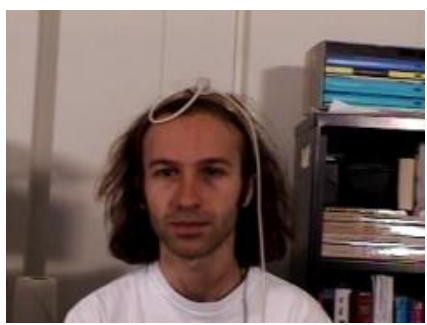 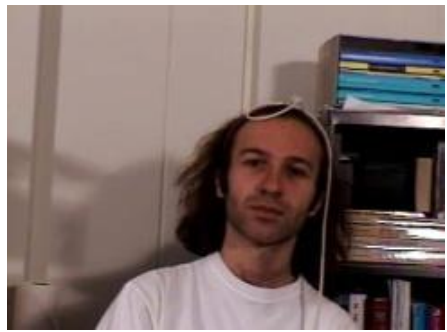 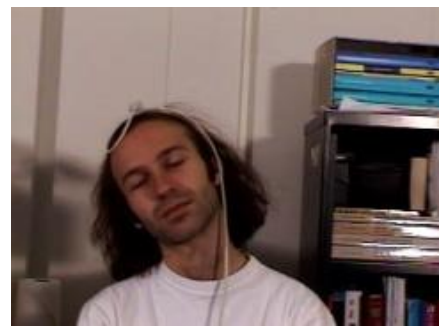

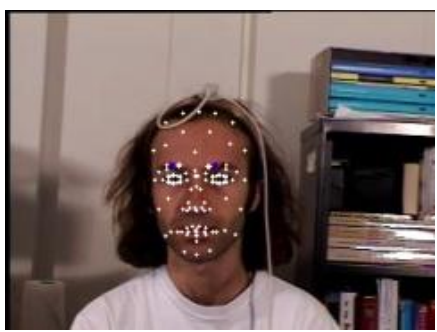 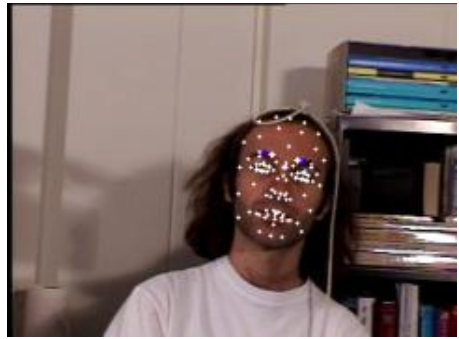 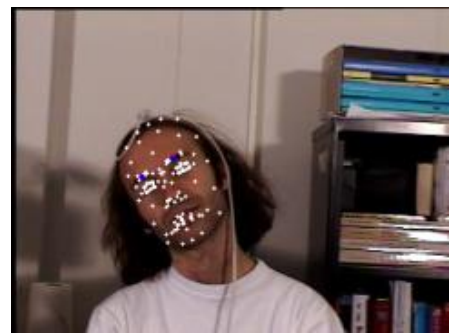

*Frame# 60*                    *Frame# 120*                    *Frame# 180*

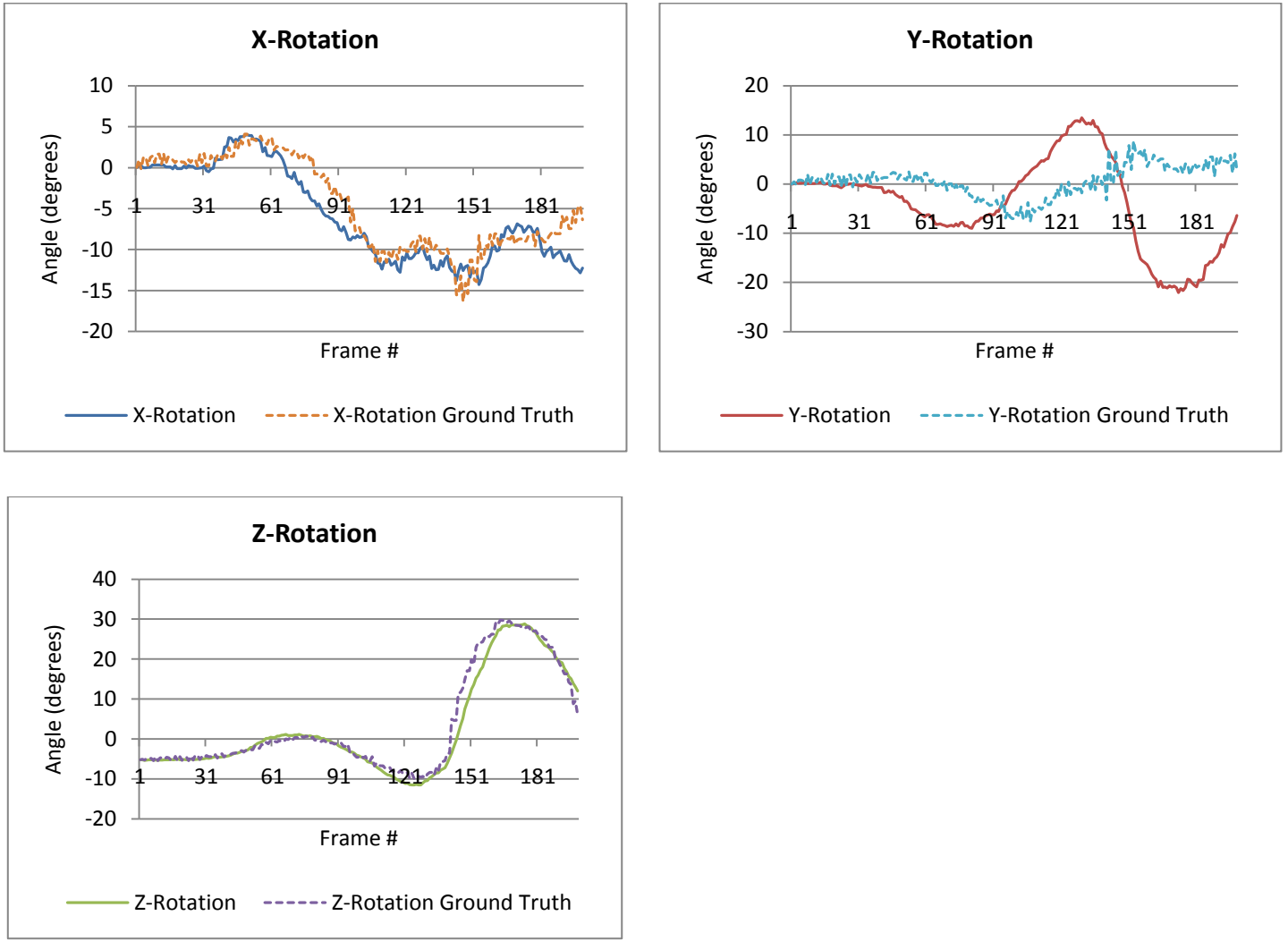

Figure 9: An example tracking sequence and its tracked result (top). In the graphs, the solid curve is our tracking result and dashed curve is the ground truth of the sequence.

We define the tracking error as follows:

$$e = \sqrt{\frac{1}{N} \sum_{i=1}^{N} \big(v(i) - g(i)\big)^2}$$

(49)

$$\overline{e} = \frac{1}{M} \sum_{j=1}^{M} e_j$$

(50)

Where $N$ is the number of frames, $M$ is the number of videos (in our case $M = 39$), $v(i)$ is the estimated value in the $i^{th}$ frame and $g(i)$ is the ground truth value in the $i^{th}$ frame. $e_{xrot}$, $e_{yrot}$ and $e_{zrot}$ are respectively the rotation errors around the x, y and z axis. In Table 3 we compare our results to the results of *Choi & Kim* [52]. *Choi & Kim* compared two of their models with the Boston University database, we will compare our results to their best results.

Table 3: The rotation errors

| Boston University data set | | | |
|---|---|---|---|
| | $\overline{e}_{xrot}$ (°) | $\overline{e}_{yrot}$ (°) | $\overline{e}_{zrot}$ (°) |
| *Our method* | 4.91 | 8.14 | 3.87 |
| *Choi & Kim* | 3.92 | 4.04 | 2.82 |

You notice that our error is relatively high compared to the result of *Choi & Kim*. The difference between our results and theirs is mainly because we do not use the same projection model. In Figure 9, you can clearly see what happens. In Figure 9, six frames from one of the test sequences are shown together with the tracking results. The graphs shows the rotation results compared to the ground truth. The thing you might notice is that the curves in the x-, and z-rotation graphs are very similar to the corresponding ground truth curves, but the curve in the y-rotation graph is not. But when you look at the tracked sequence you see that the head is correctly tracked. This is caused by our projection model. Our tracker perceives the perspective change caused by for example translation, as a rotational change. Visually, this won't make a difference, but when comparing the results, it will introduce some additional error. Despite of this, it is still interesting to compare the results with the Boston University ground truth.

There is one video where our tracker lost track of the head. In this video, our tracker lost track twice, but it recovered itself very well and was able to carry on with the tracking process. Figure 10 and Figure 11 show the erroneous tracking and the result after recovery. The graphs show clearly that the tracker loses the head at frame 69 and 136 and recovers by resetting the tracker. Once the tracker resets, the tracker immediately tries to find a new and better estimate.

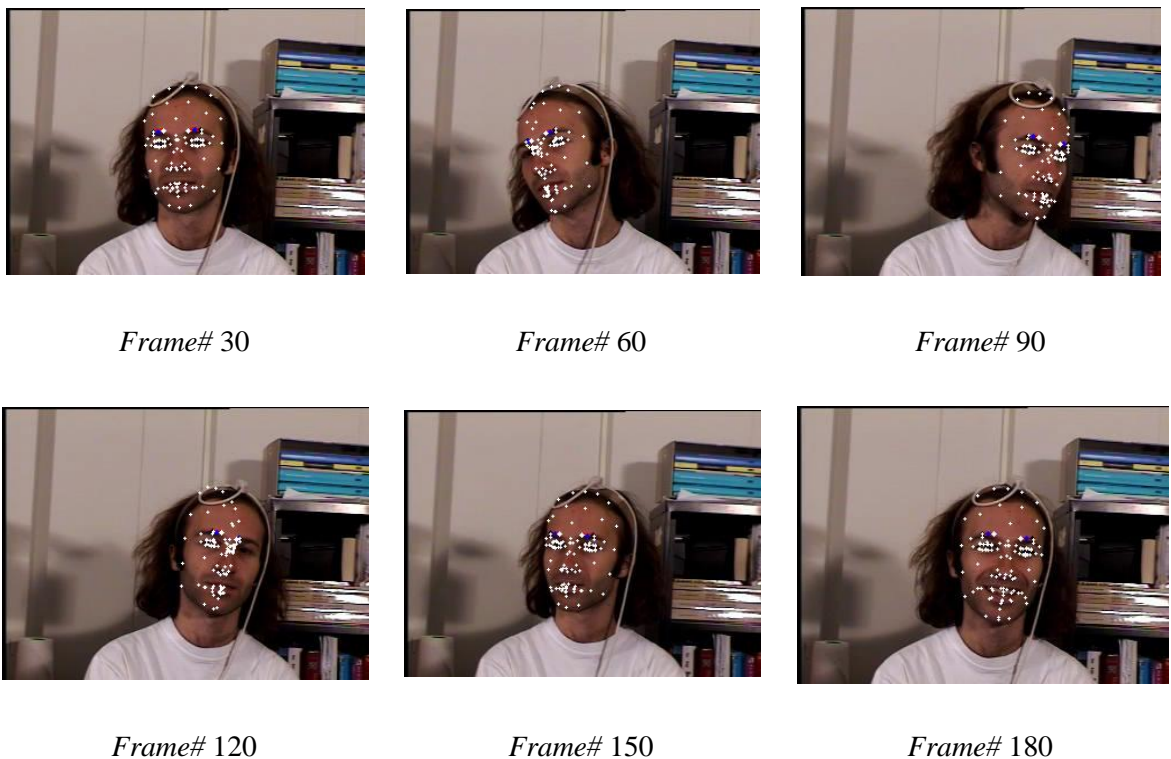

*Frame#* 30          *Frame#* 60          *Frame#* 90

*Frame#* 120          *Frame#* 150          *Frame#* 180

Figure 10: The video sequence where the tracker lost the head for a short moment.

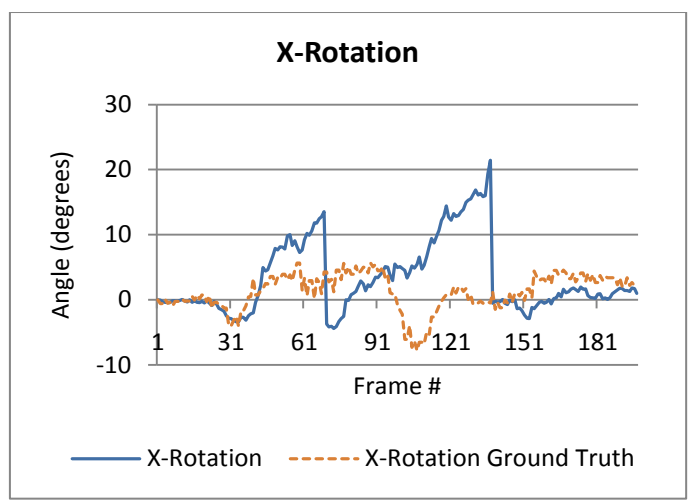

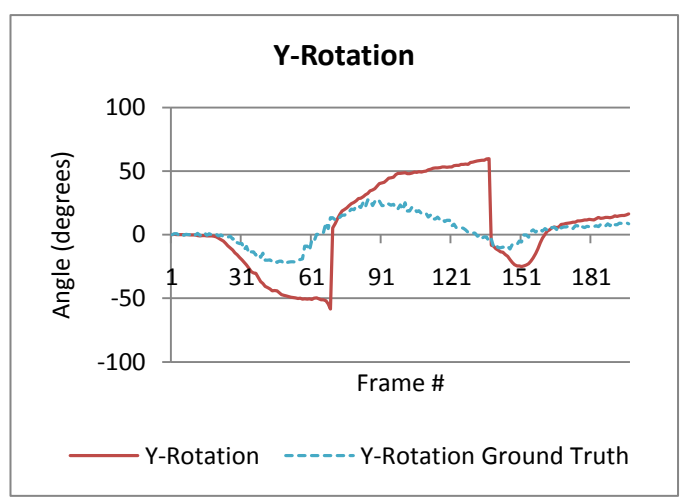

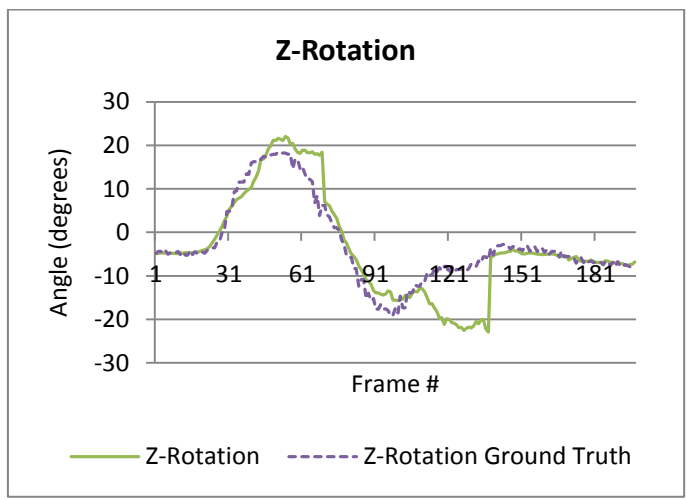

Figure 11: The graphs clearly show where the tracker lost the head and how it has recovered and continues tracking.

## 5.1.2 Experiment 2: Translation and Rotation

This experiment is designed to evaluate both the translational and rotational capabilities of the tracker. Here we don't use the Boston University data set, but our own captured videos with an semi-manual annotated ground truth. We have acquired the ground truth data by fine-tuning the pose estimate of the tracker. The videos are captured using a Logitech Quickcam 9000 webcam and in an uniformly lit room. The videos have a resolution of 640x480 and a frame rate of 30 FPS. In this experiment we use twelve videos in total with two different subjects, a male (*Vid1-Vid6*) and a female (*Vid7-Vid12*). For each subject we captured three translation and three rotation videos. One video of each subject is shown in Figure 12. The maximum accuracy of the ground truth is 1.0 pixel for translation and scaling, and 1.0 degree for the rotation parameters. The downside of annotating the data by hand is that it is not very precise. It is very difficult to give a precise estimate of the orientation of the head. But because we are interested in acquiring believable estimation of the position and orientation and not in a precise error measurement, it is not a problem.

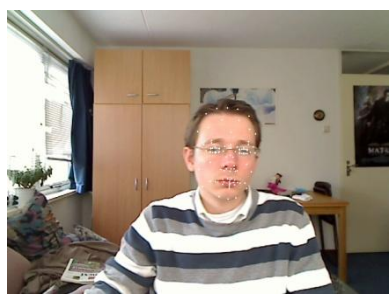 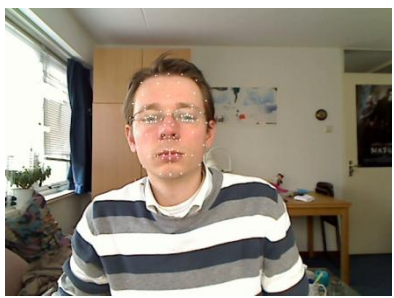 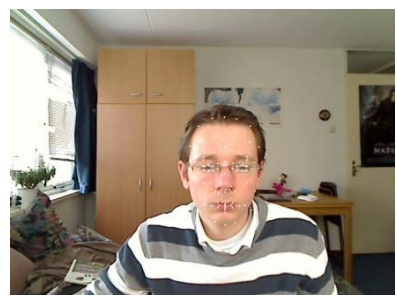

*Frame#* 64          *Frame#* 128          *Frame#* 196

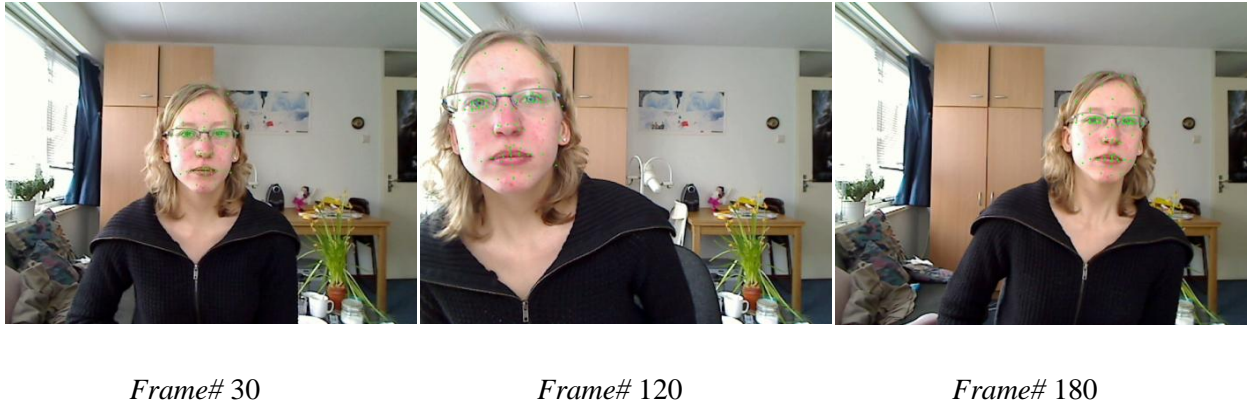

|              |              |              |
| :----------: | :----------: | :----------: |
| *Frame#* 30  | *Frame#* 120 | *Frame#* 180 |

Figure 12: Two translation video sequences. The top sequence shows the male subject and the bottom sequence shows the female subject.

Table 4 and Table 5 show the tracking results respectively from the translation and the rotation videos. The results from Table 4 show that tracking translational movements can be done with a small error. Rotations are more difficult for our tracker and therefore results in higher error values. Rotations might be more difficult for our tracker, but the errors in Table 5 are still small. Also from looking at Figure 13 you see that the tracker can sometimes be slightly off, but it still follows the general curve very well. When you compare the results of the male (*Vid1-Vid6*) and female videos (*Vid7-Vid12*), you notice that there is not very much difference in accuracy between them. Especially the translational results are very close to each other. The differences between the rotation errors of the male and female videos are small.

Table 4: Tracking results from the translation videos

| Translation video results | | | | | |
| :---: | :---: | :---: | :---: | :---: | :---: |
| $e_{xtrans}$ (*pixels*) | $e_{ytrans}$ (*pixels*) | $e_{scale}$ | $e_{xrot}$ (°) | $e_{yrot}$ (°) | $e_{zrot}$ (°) |

| | $e_{xtrans}$ (*pixels*) | $e_{ytrans}$ (*pixels*) | $e_{scale}$ | $e_{xrot}$ (°) | $e_{yrot}$ (°) | $e_{zrot}$ (°) |
| :---: | :---: | :---: | :---: | :---: | :---: | :---: |
| *Vid1* | 0,12 | 0,11 | 0,38 | 0,95 | 0,63 | 0,19 |
| *Vid2* | 0,64 | 0,47 | 1,47 | 2,06 | 2,21 | 1,08 |
| *Vid3* | 0,19 | 0,24 | 0,45 | 1,14 | 0,51 | 0,24 |
| *Vid7* | 0,15 | 0,37 | 0,41 | 2,39 | 0,82 | 0,33 |
| *Vid8* | 0,44 | 0,55 | 0,80 | 1,48 | 1,15 | 0,52 |
| *Vid9* | 0,09 | 0,08 | 0,16 | 0,66 | 0,53 | 0,18 |
| *Average* | 0,27 | 0,30 | 0,61 | 1,45 | 0,98 | 0,42 |

Table 5: Tracking results from the rotation videos. The highlighted cells contain of the error values associated with the the main rotation of the video.

| | $e_{xtrans}$ (pixels) | $e_{ytrans}$ (pixels) | $e_{scale}$ (pixels) | $e_{xrot}$ (°) | $e_{yrot}$ (°) | $e_{zrot}$ (°) |
|---|---|---|---|---|---|---|
| | | | Rotation video results | | | |
| *Vid4 (X rot.)* | 0,17 | 0,27 | 0,55 | 3,68 | 1,42 | 0,83 |
| *Vid5 (Y rot.)* | 0,80 | 0,56 | 1,13 | 1,95 | 4,68 | 1,18 |
| *Vid6 (Z rot.)* | 0,69 | 0,52 | 3,04 | 3,13 | 2,93 | 0,95 |
| *Vid10 (X rot.)* | 0,47 | 0,27 | 0,71 | 3,59 | 1,25 | 0,65 |
| *Vid11 (Y rot.)* | 0,52 | 0,17 | 0,44 | 1,20 | 1,72 | 0,73 |
| *Vid12 (Z rot.)* | 0,30 | 0,31 | 1,42 | 3,14 | 2,10 | 1,68 |
| *Average* | 0,49 | 0,35 | 1,22 | 2,78 | 2,35 | 1,00 |

We must keep in mind that the ground truth that is used to calculate the error does not have a perfect precision. To obtain the ground truth we only corrected the tracker when it was noticeable visually off. The ground truth is thus very much based on the tracking results of our tracker. This means that when our tracker is not noticeable off, the ground truth value is very close to the tracked result. Nevertheless, the numerical and visual results are good, and the tracker has never lost the head. Also, the results show that there is no significant difference in performance when using different subjects.

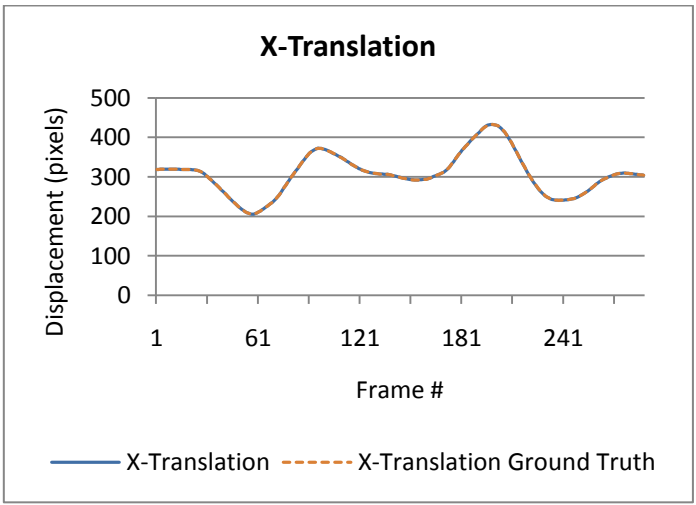

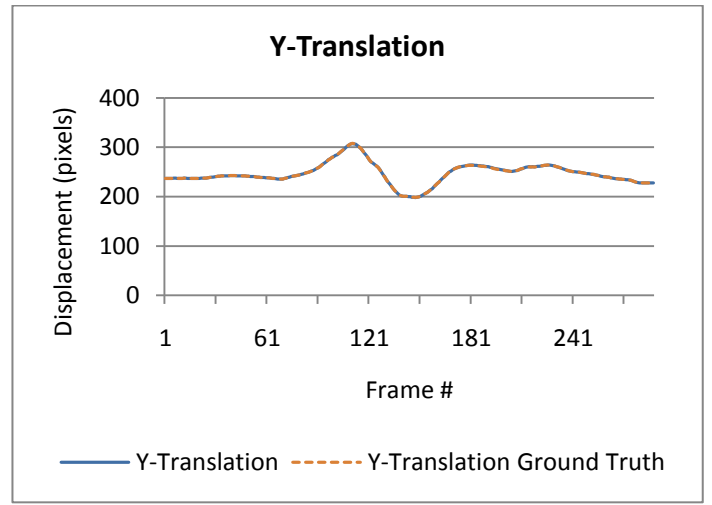

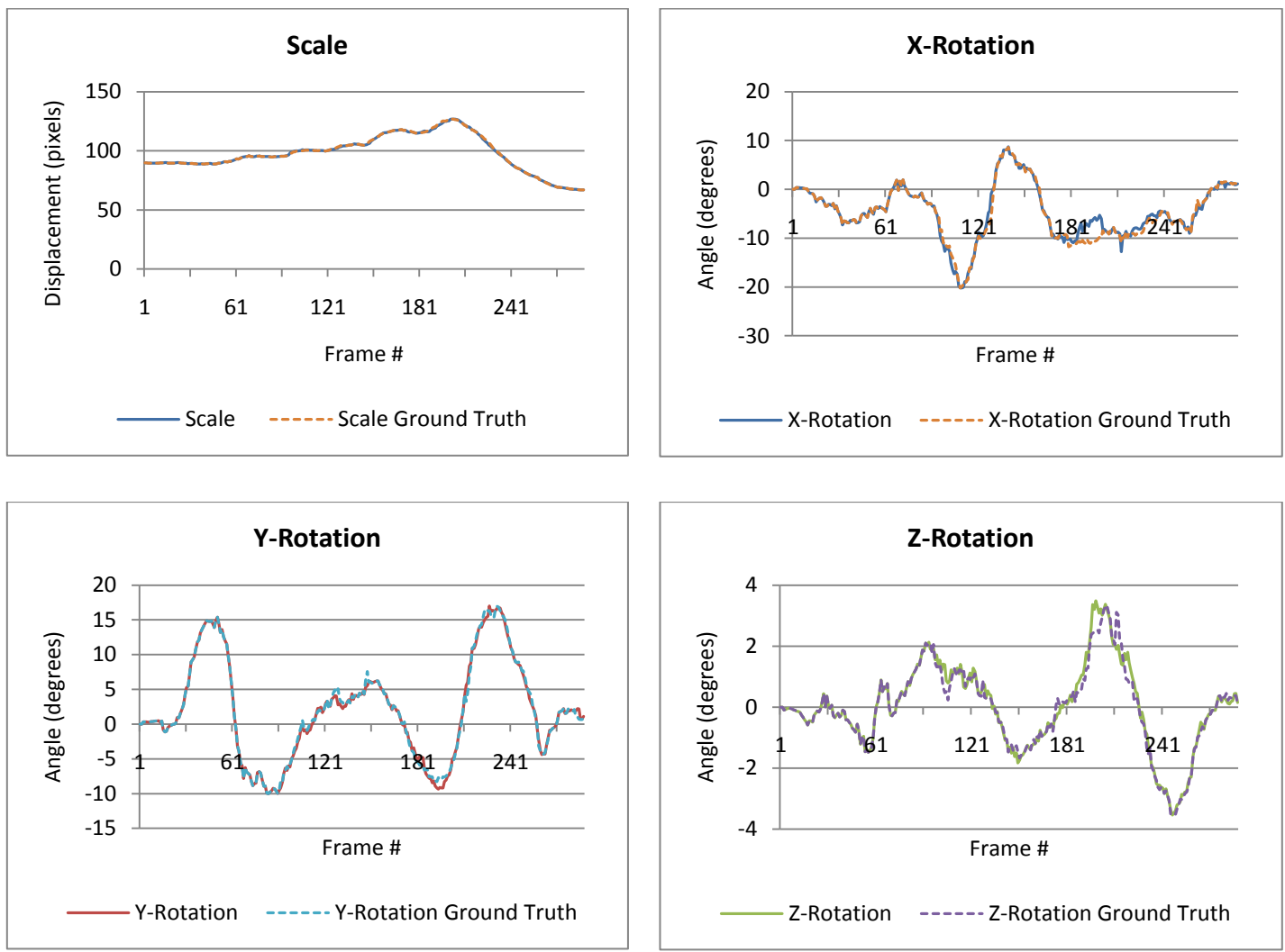

Figure 13: Translation and rotation graphs of the (male) video, vid3.

### 5.1.3 Experiment 2: Fast movement

In this experiment we evaluate the robustness of our tracker when dealing with fast moving subjects. We also evaluate how much influence our optical flow pre-estimation has on the accuracy of the tracker. This test set consists of twelve video sequences with male and female subjects. For each subject there are six videos with each one containing one dominating head movement; horizontal translation, vertical translation, scaling, x-rotation, y-rotation and z-rotation. All the movements in these videos are fast i.e. the image gets blurred because of the fast moving subject (see Figure 14, frame 65). The performance of our tracker with and without the optical flow pre-estimation (NOF: no optical flow pre-estimation) is compared with the ground truth.

Table 6 and Table 7 show the error values respectively of the six male and six female videos with their different tracker configurations. As you can see, tracking a fast moving subject does increase the tracking error, but the tracker result is still very acceptable. Also, using the optical flow pre-estimation for the template tracker reduces the error significantly. Optical flow pre-estimation improves the average total rotational error ($e_{zrot}$) for these six male videos with a factor 2.0 and for the female videos with a factor 2.7. The average total translation error ($e_{trans}$) is even improved with a factor 5.5 for the male videos and with a factor 9.4 for the female videos. This is a big improvement and especially noticeable by the reduction of the number of head losses. Although there is a big difference in performance, most of the performance increase is when moving vertically or horizontally. With fast head rotation, the difference between pre-estimation and no pre-estimation is not that big. So, fast rotation does not benefit that much of the pre-estimation. A reason for this is that, when the head rotates, there is not very much point pixel movement in the image plane. Most of the movement will take place in the z-direction (except when rotating around the z-axis) and is therefore not very noticeable in the image.

We should again mention that we acquired the ground truth data by correcting the estimation from the tracker. This means that it is more likely that the ground truth is closer to the tracker estimate with optical flow pre-estimation then without.

Table 6: All the error values of the different male videos. The highlighted cells contain the error values associated with the the main movement of the video.

| | $e_{xrot}$ (°) | $e_{yrot}$ (°) | $e_{zrot}$ (°) | $e_{rot}$ (°) | $e_{scale}$ (pixels) | $e_{xtrans}$ (pixels) | $e_{ytrans}$ (pixels) | $e_{trans}$ (pixels) |
|---|---|---|---|---|---|---|---|---|
| | | | | Fast male movement video results | | | | |
| *Hor. Vid* | 1,41 | 1,59 | 0,39 | 3,40 | 0,65 | 0,75 | 0,31 | 1,72 |
| *Hor. Vid NOF* | 2,95 | 3,20 | 1,13 | 7,28 | 1,41 | 6,18 | 0,78 | 8,37 |
| *Vert. Vid* | 5,10 | 1,44 | 0,99 | 7,52 | 0,98 | 0,22 | 1,02 | 2,23 |
| *Vert. Vid NOF* | 18,30 | 9,95 | 0,72 | 28,97 | 3,05 | 5,56 | 24,12 | 32,72 |
| *Scaling Vid* | 0,00 | 0,00 | 0,00 | 0,00 | 0,00 | 0,00 | 0,00 | 0,00 |
| *Scaling Vid NOF* | 1,59 | 0,72 | 0,30 | 2,60 | 0,48 | 0,36 | 0,47 | 1,32 |
| *X-rot. Vid* | 5,66 | 2,29 | 0,81 | 8,76 | 1,50 | 0,52 | 0,59 | 2,61 |
| *X-rot. Vid NOF* | 7,05 | 3,91 | 1,57 | 12,52 | 2,89 | 0,83 | 8,36 | 12,08 |
| *Y-rot. Vid* | 2,15 | 3,56 | 1,19 | 6,91 | 0,66 | 1,03 | 0,46 | 2,14 |
| *Y-rot. Vid NOF* | 2,92 | 4,46 | 1,81 | 9,20 | 2,36 | 1,66 | 0,72 | 4,74 |
| *Z-rot. Vid* | 1,62 | 5,44 | 1,99 | 9,05 | 0,77 | 1,16 | 0,98 | 2,91 |
| *Z-rot. Vid NOF* | 3,06 | 4,93 | 3,05 | 11,05 | 0,79 | 1,93 | 1,70 | 4,42 |
| *Average* | 2,66 | 2,39 | 0,90 | 5,94 | 0,76 | 0,61 | 0,56 | 1,93 |
| *Average NOF* | 5,98 | 4,53 | 1,43 | 11,94 | 1,83 | 2,75 | 6,03 | 10,61 |

Table 7: All the error values of the different female videos. The highlighted cells contain the error values associated with the the main movement of the video.

| | $e_{xrot}$ (°) | $e_{yrot}$ (°) | $e_{zrot}$ (°) | $e_{rot}$ (°) | $e_{scale}$ (pixels) | $e_{xtrans}$ (pixels) | $e_{ytrans}$ (pixels) | $e_{trans}$ (pixels) |
|---|---|---|---|---|---|---|---|---|
| | | | Fast female movement video results | | | | | |
| *Hor. Vid* | 1,82 | 1,55 | 0,69 | 4,06 | 0,45 | 0,58 | 0,34 | 1,36 |
| *Hor. Vid NOF* | 3,57 | 10,37 | 3,65 | 17,58 | 9,14 | 28,75 | 27,87 | 65,76 |
| *Vert. Vid* | 1,33 | 0,99 | 0,44 | 2,76 | 0,57 | 0,23 | 0,25 | 1,05 |
| *Vert. Vid NOF* | 9,66 | 3,21 | 0,86 | 13,73 | 1,01 | 1,03 | 4,03 | 6,06 |
| *Scaling Vid* | 3,37 | 1,30 | 0,77 | 5,45 | 0,81 | 0,72 | 0,67 | 2,20 |
| *Scaling Vid NOF* | 13,00 | 18,67 | 2,20 | 33,87 | 5,90 | 9,07 | 2,43 | 17,39 |
| *X-rot. Vid* | 3,09 | 0,78 | 0,36 | 4,23 | 0,44 | 0,16 | 0,20 | 0,81 |
| *X-rot. Vid NOF* | 3,43 | 0,95 | 0,47 | 4,85 | 0,50 | 0,23 | 0,47 | 1,20 |
| *Y-rot. Vid* | 3,54 | 3,09 | 0,95 | 7,58 | 1,01 | 0,88 | 0,97 | 2,87 |
| *Y-rot. Vid NOF* | 2,59 | 3,08 | 1,52 | 7,19 | 1,45 | 1,81 | 0,64 | 3,90 |
| *Z-rot. Vid* | 3,14 | 2,10 | 1,68 | 6,93 | 1,42 | 0,30 | 0,31 | 2,04 |
| *Z-rot. Vid NOF* | 2,91 | 2,01 | 2,01 | 6,93 | 1,28 | 0,53 | 0,56 | 2,37 |
| *Average* | 2,72 | 1,63 | 0,81 | 5,17 | 0,78 | 0,48 | 0,46 | 1,72 |
| *Average NOF* | 5,86 | 6,38 | 1,78 | 14,03 | 3,21 | 6,90 | 6,00 | 16,12 |

In Figure 14a, a number of frames from vertical movement test sequence are shown. Here you see that the tracker with the optical flow pre-estimation is able to track the fast vertical head movement and the tracker without the pre-estimation (Figure 14b) is not.

One of the main problems of fast movement is that the images get blurred and therefore important information is lost. When there is not enough information, the template tracker won't perform as well as it should. This will lead to poor performance and erroneous tracking.

We can conclude from the experiment results that the robustness against fast moving subjects can be effectively increased by using the optical flow pre-estimation as initial estimate for the template tracker. Using the optical flow pre-estimation reduces the number of head losses significantly.

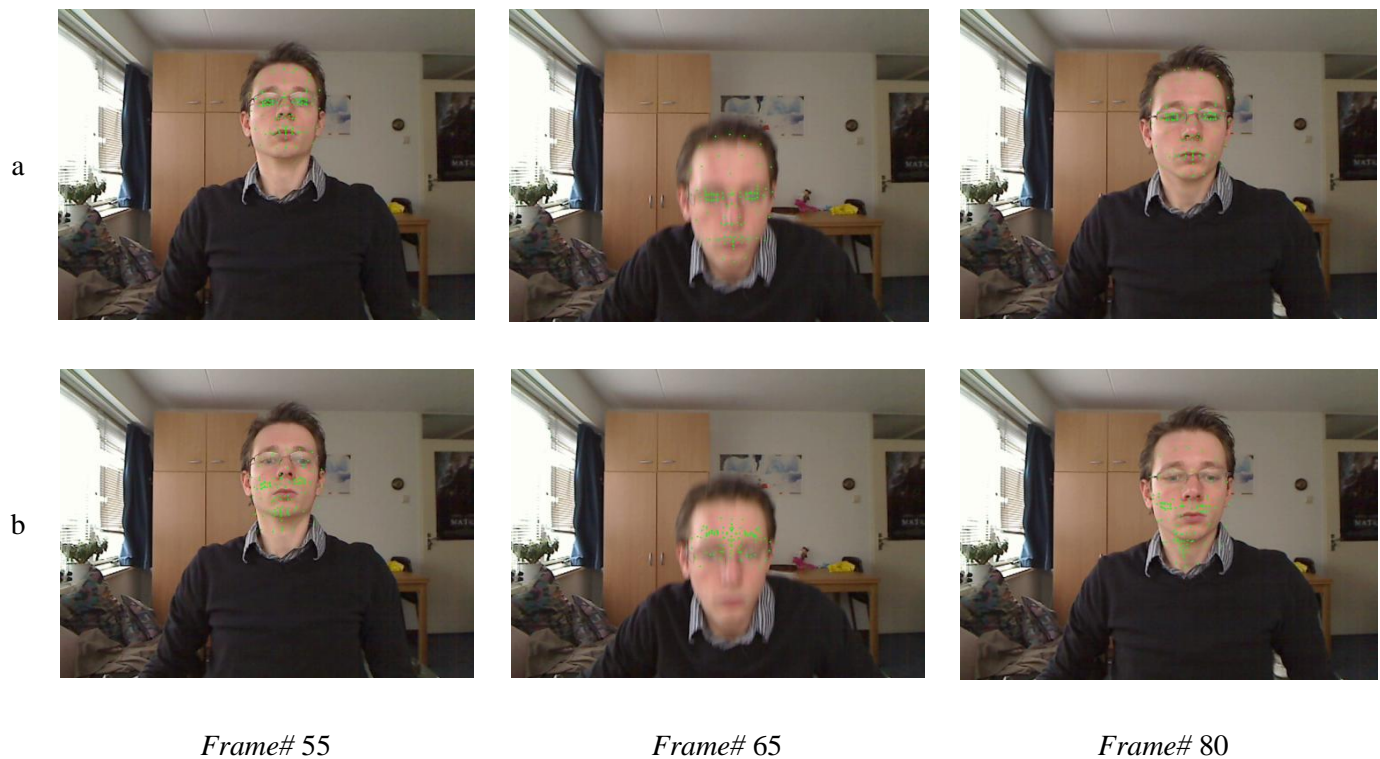

<div align="center">Frame# 55           Frame# 65           Frame# 80</div>

Figure 14:Fast vertical movement (male) video sequence (*Vert. Vid*). a) The tracking results when using the optical flow pre-estimation. b) The tracking results when only using the template tracking method.

### 5.1.4 Experiment 3: Occlusion

This experiment is designed to measure how robust our tracker is against partial occlusions of the subject. We have recorded a number of video sequences with different kinds of occlusions, and measured the performance of our tracker. We also measure the effectiveness of the outlier detection and removal method.

We use six video sequences with different partial occlusions and movements. In the videos, the head is occluded by a hand, an orange box or a white square. A screenshot of each video is shown in Figure 15.

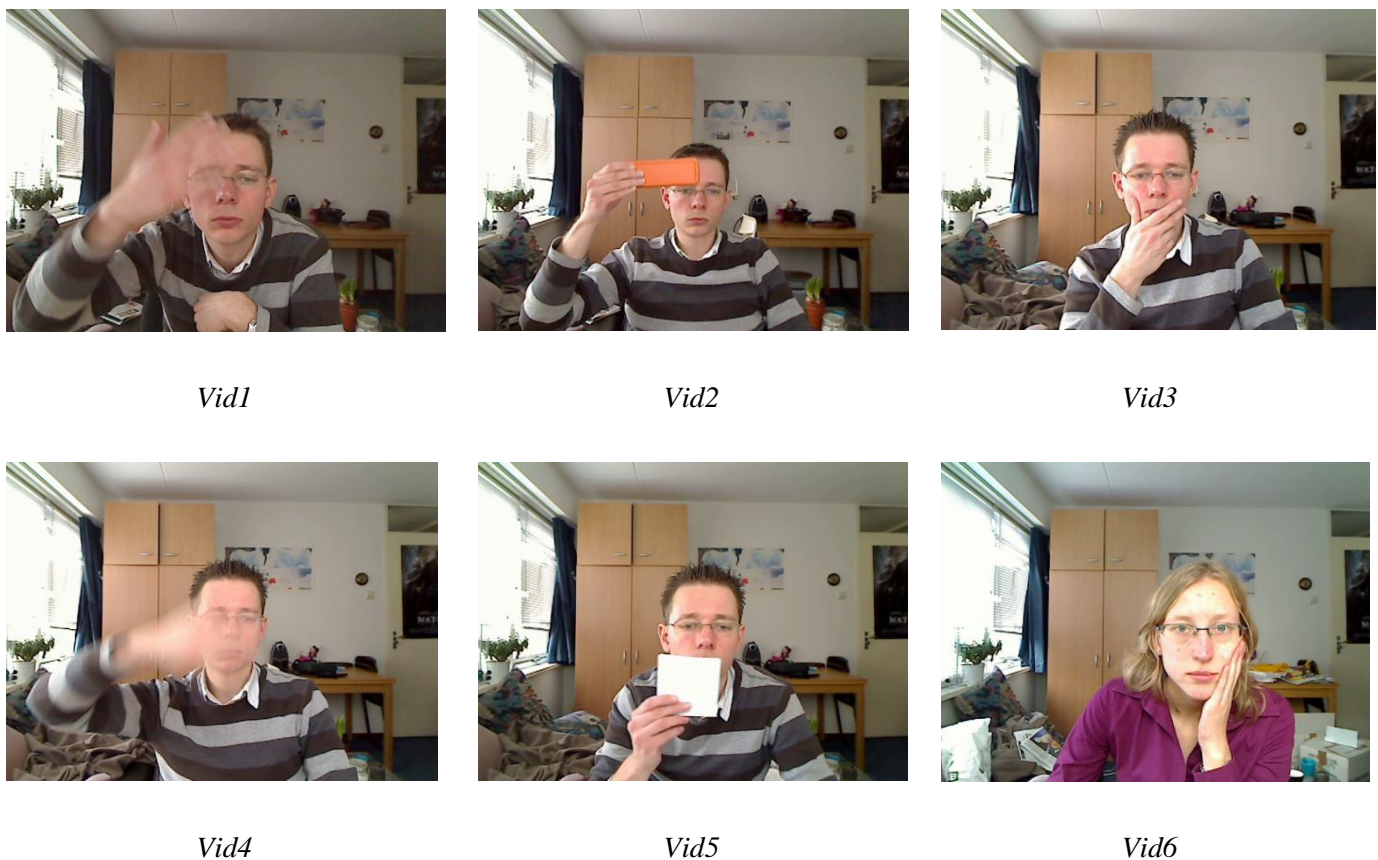

*Vid1*        *Vid2*        *Vid3*

*Vid4*        *Vid5*        *Vid6*

Figure 15: Screenshots of the six videos used in this partial occlusion experiment.

Table 8 summarizes our tracking results. The results show that our tracker can handle small partial occlusions and has a relative small error. But we noticed that larger occlusions are a problem and results in losing the head. To further investigate the influence of our outlier removal on the results, we compared the error values with the results of our tracker without outlier removal. These results are shown in Table 9. Table 9 summarizes the difference in error between the tracker with outlier removal and without outlier removal. As you can see, using outlier removal does reduce the error values in most of the videos, but unfortunately, when visually inspecting the results, outlier removal does not allow more occlusion. If too much of the head is occluded than our tracker will fail and tries to reinitialize the tracker. Outlier removal maybe has not a really big effect on coping with occlusions, but it has an effect on the stability and robustness of the tracker i.e. it will be less likely that the tracker will lose the head due to the partial occlusion. This is because the stability of the tracker greatly depends on the error of the template tracker (see equation (30)). When we examine the error values of purely the template tracker, we notice that using the outlier removal significantly reduces the influence of the occlusion on this template error.

Table 8: The error values of the different videos with partial occlusion.

| | Error values | | | | | | | |
|---|---|---|---|---|---|---|---|---|
| | $e_{xrot}$ (°) | $e_{yrot}$ (°) | $e_{zrot}$ (°) | $e_{rot}$ (°) | $e_{scale}$ (pixels) | $e_{xtrans}$ (pixels) | $e_{ytrans}$ (pixels) | $e_{trans}$ (pixels) |
| Vid1 | 1,55 | 2,42 | 0,33 | 4,30 | 0,57 | 0,19 | 0,45 | 1,22 |
| Vid2 | 0,70 | 0,49 | 0,26 | 1,45 | 0,22 | 0,19 | 0,09 | 0,51 |
| Vid3 | 0,69 | 0,64 | 0,24 | 1,57 | 0,28 | 0,16 | 0,10 | 0,54 |
| Vid4 | 1,75 | 1,37 | 1,91 | 5,04 | 0,65 | 0,68 | 0,81 | 2,13 |
| Vid5 | 2,17 | 1,77 | 0,31 | 4,26 | 0,43 | 0,16 | 0,12 | 0,71 |
| Vid6 | 0,36 | 0,66 | 0,22 | 1,24 | 0,09 | 0,10 | 0,06 | 0,25 |
| Average | 1,20 | 1,23 | 0,55 | 2,97 | 0,37 | 0,25 | 0,27 | 0,89 |

Table 9: The rotational and translational error difference between tracking with and without outlier removal. Positive values means an improvement. The large value of *Vid5* is because the tracker lost the head.

| | Outlier removal comparison | |
|---|---|---|
| | $de_{rot}$ (°) | $de_{trans}$ (pixels) |
| Vid1 | -2,48 | -0,36 |
| Vid2 | 1,19 | 0,12 |
| Vid3 | 6,19 | 1,22 |
| Vid4 | 0,94 | 0,11 |
| Vid5 | 1,99 | 124,04 |
| Vid6 | 1,96 | 0,86 |
| Average | 1,63 | 21,00 |

In Figure 16, the different template tracker error values are plotted. As you can see, it is clear that removing the outliers reduces the template tracker error significantly. This means that outliers (occluded parts) are identified correctly and removed, which results in a more stable tracker and less head losses.

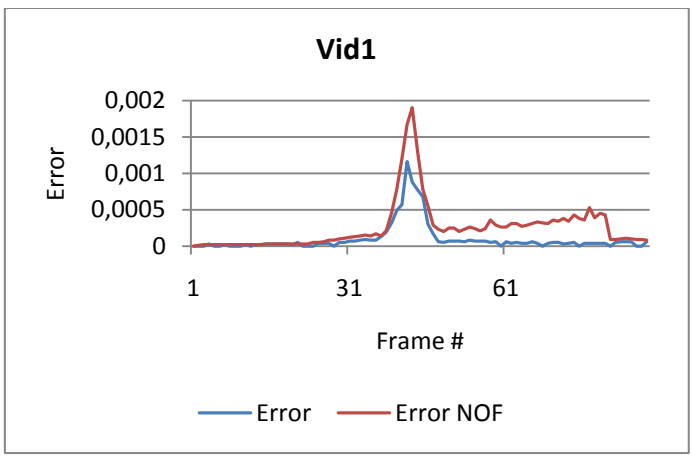

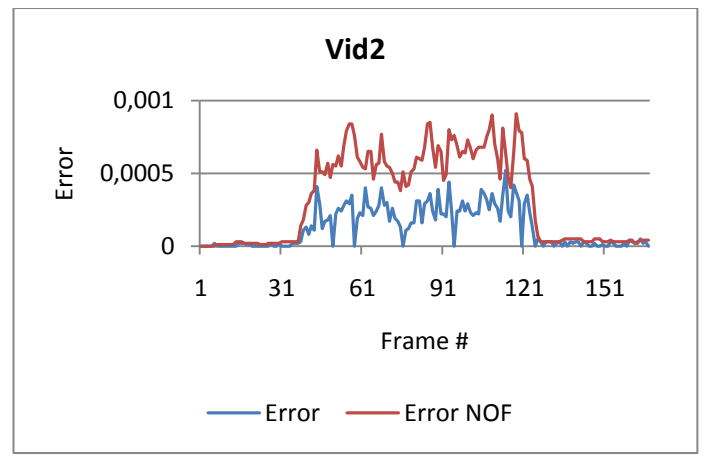

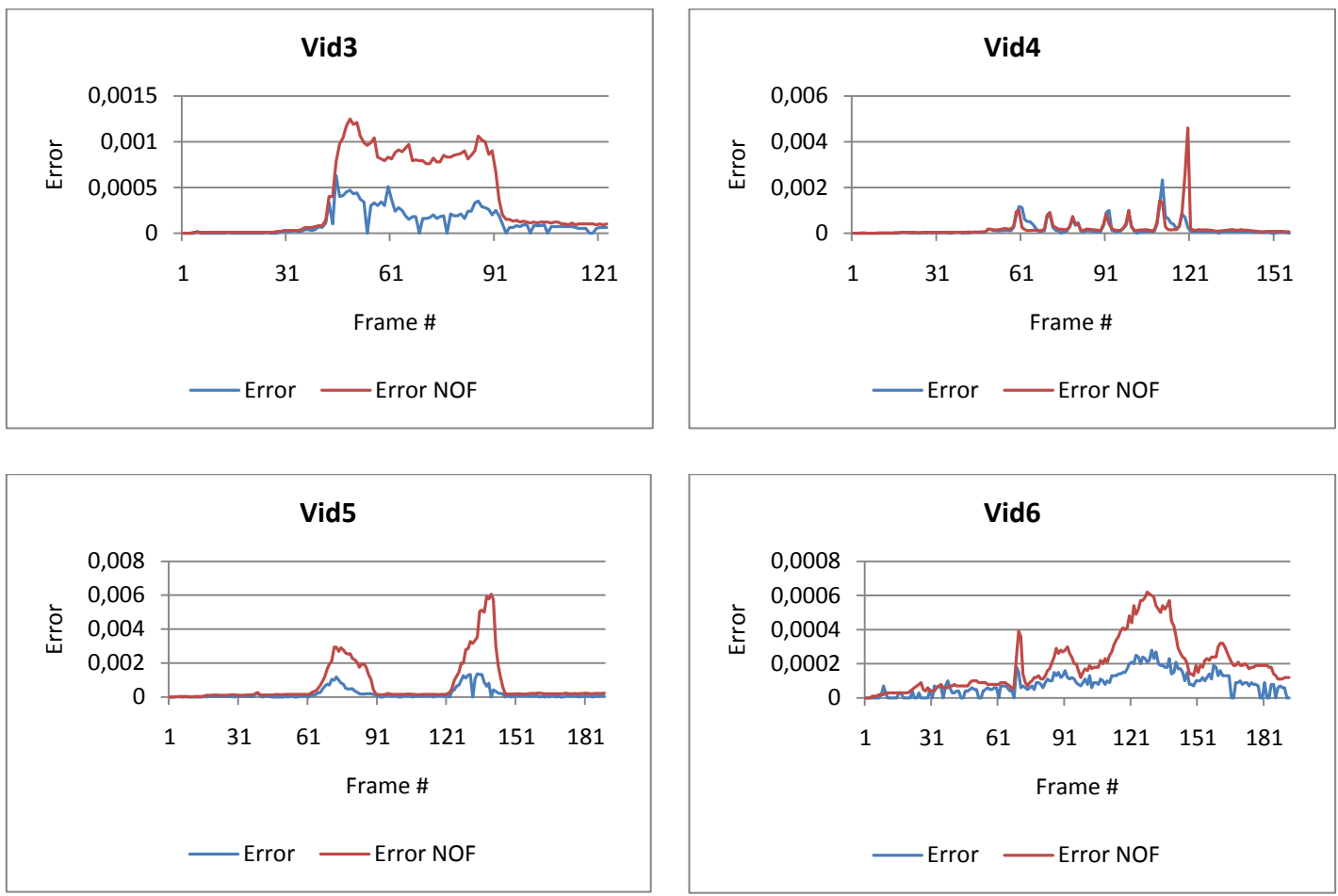

Figure 16: The template error values of the different videos. The blue curve denotes the error of the tracker with outlier removal and the red curve denotes the error values without outlier removal.

In summary, our tracker is robust against small partial occlusions, but if a large area or if important (information rich) parts are occluded, our tracker will probably fail. Luckily, the tracker can automatically restart the tracking process when the occlusion is gone. The main benefit of the outlier removal is that is reduces the number of head losses caused by occlusion.

## 5.2 Mouth and Eyebrow tracking

To evaluate the performance of the mouth and eyebrow tracker we conduct the same kind of experiments as in the previous sections. We have captured a number of videos where the subject performs a variety of facial expressions. To evaluate the performance, we visually inspect the videos and the produced results.

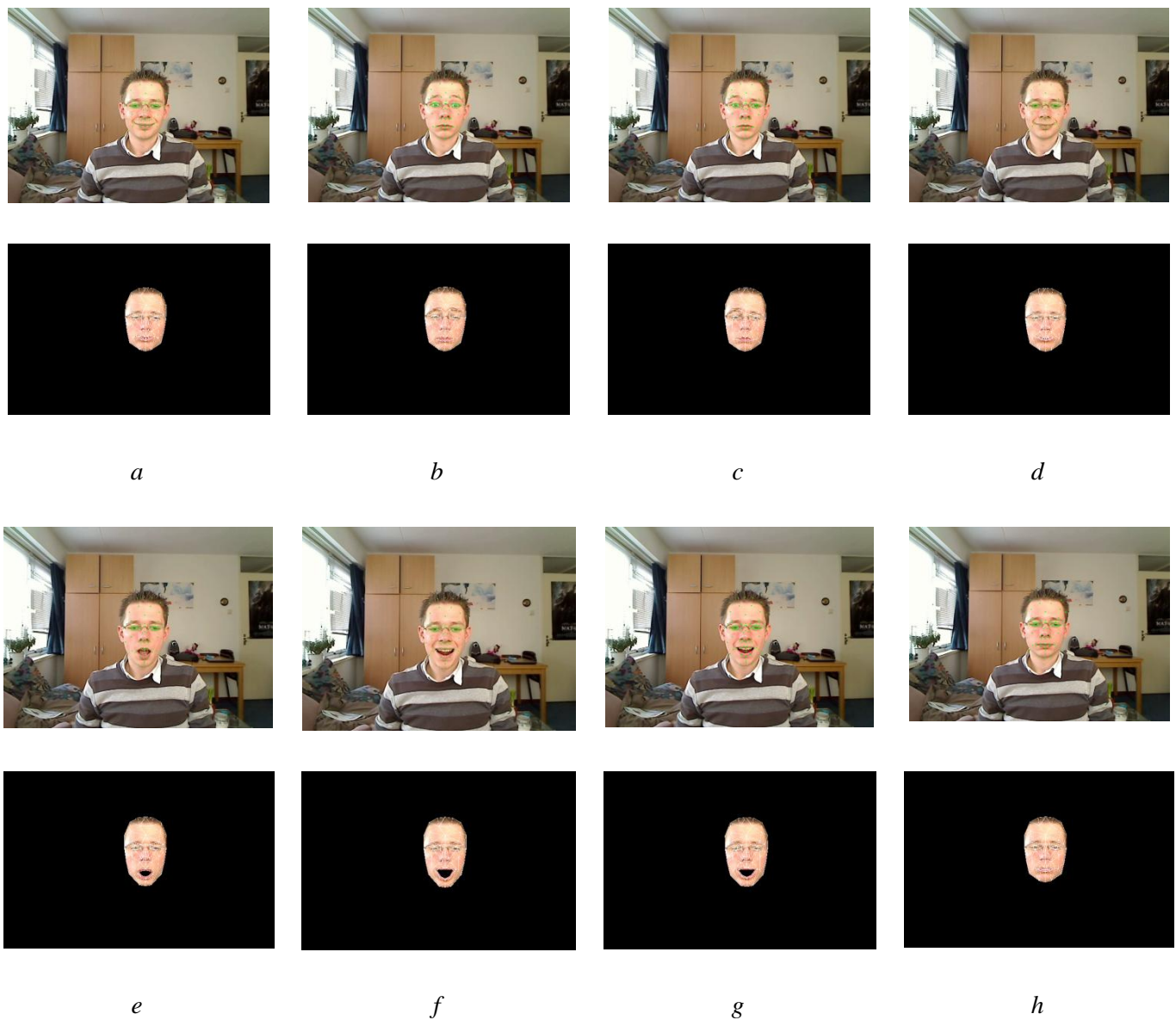

*a*                    *b*                    *c*                    *d*

*e*                    *f*                    *g*                    *h*

Figure 17: Expression video 1, where you see the expression video and the deformed 3D model representing the expression.

As you can see in Figure 17, the tracker is capable of tracking the facial expression in this video. It tracks the both the eyebrows (Figure 17, b and c) and the mouth expressions. In Figure 18 and Figure 19 the results of going from a neutral face to three basis facial expressions are shown. You can easily identify the expression from only looking at the 3D model image. As you can see, there is not much difference between the *happy* and the *scared* expression. Sometimes the produced results are not very distinctive, this is because the tracker can only do four different mouth deformations and thus cannot capture all the subtle changes. Increasing the number of model deformations will probably give a more accurate expression, but this will decrease the stability. Nevertheless, you can identify the expression obtained by the tracker.

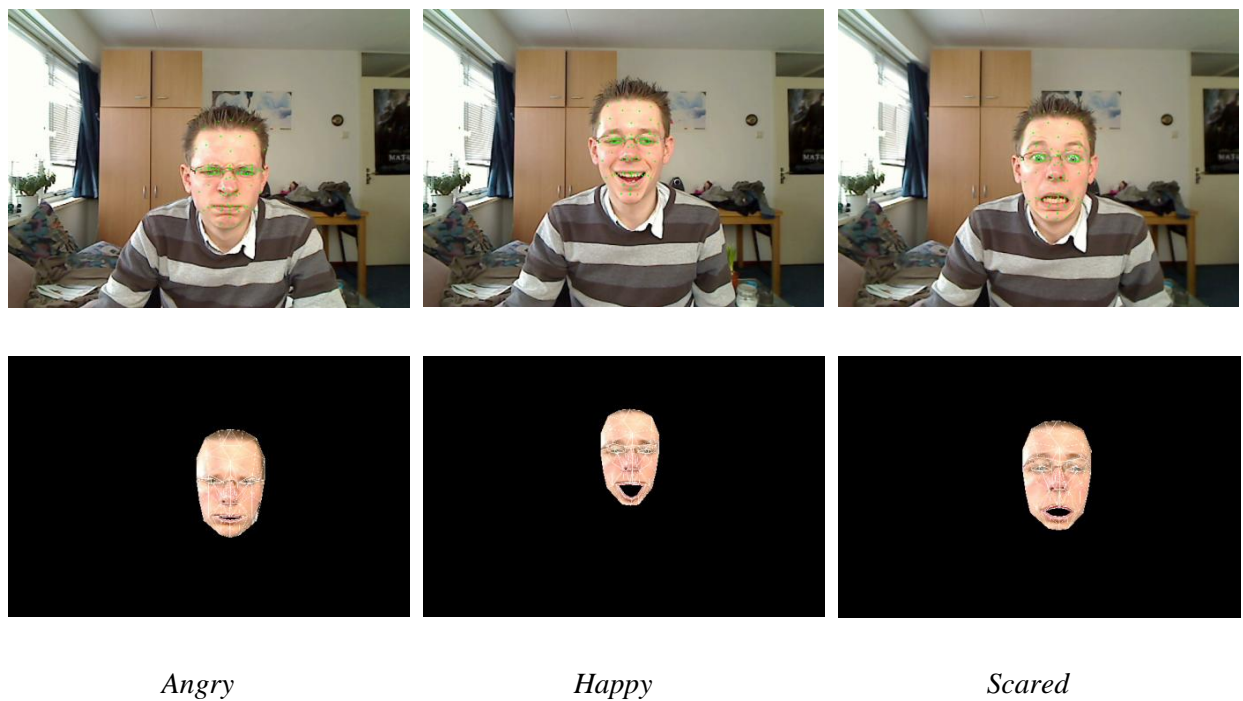

*Angry*                    *Happy*                    *Scared*

Figure 18: Three facial expressions of a male subject and the tracking results.

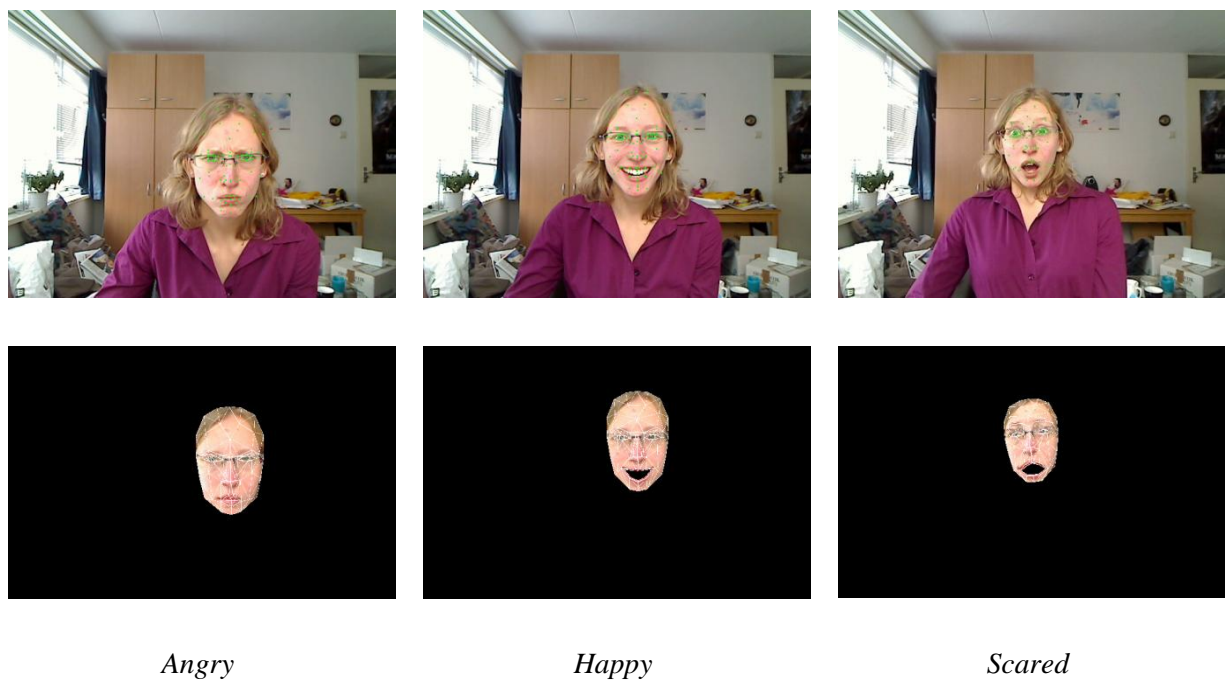

*Angry*                    *Happy*                    *Scared*

Figure 19: Three facial expressions of a female subject and the tracking results.

There is a chance that the mouth tracker looses track or gives a wrong result like is shown in Figure 20. This happens when the expression is too complex for our tracker, if the head is occluded or when the head estimate is not precise enough. Fortunately, the mouth tracker is capable of recovering itself like shown in Figure 20. It is also possible for the eyebrow tracker to get stuck in the wrong 'place'.

This sometimes happens when the head estimate was slightly off. But because the eyebrows are not as complex as the mouth, they are much more stable and recover quickly.

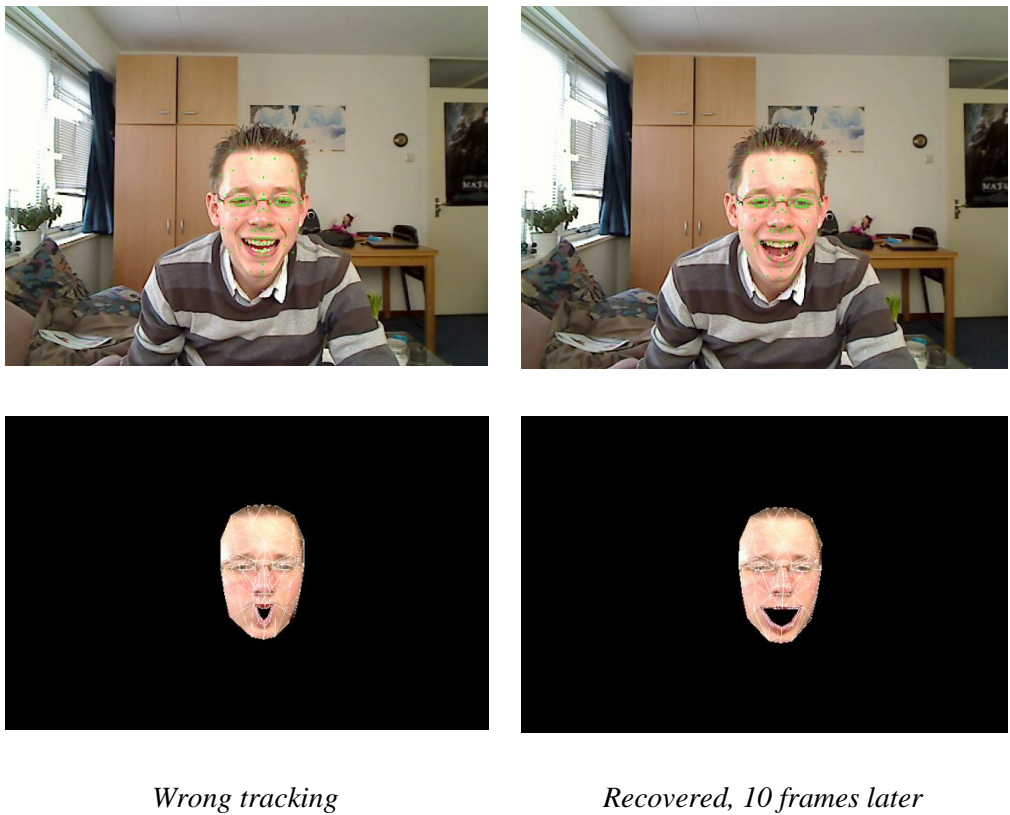

*Wrong tracking*                    *Recovered, 10 frames later*

Figure 20: Example of wrongly tracked mouth and its recovery ten frames later.

The combination of our eyebrow and mouth tracker makes it possible to track facial expressions. The mouth and eyebrow trackers can successfully track the features so that basic expressions can be identified. Because of the complexity of the mouth and eyebrows, it is possible that the mouth and eyebrow trackers are off, but when this happens, it often recovers itself quickly.

## 5.3   Computation Time and Real-time Performance

We have measured the total and individual computation time of each component of the framework i.e. optical flow pre-estimation, template tracking, eyebrow tracking and mouth tracking. We use 33 head and expression tracking video sequences to compute the average computation time. Table 10 and Figure 21 show the computation time (in milliseconds) of each individual component.

Table 10: Average computation time (in milliseconds) of each component. The average computation time is calculated using 33 different videos.

| | Computation time (ms) | | | | | |
|---|---|---|---|---|---|---|
| | *Opt. Flow pre-estimation* | *Template tracker* | *Head tracker total* | *Eyebrow tracker* | *Mouth tracker* | *Total* |
| *Translation* | 18 | 13 | 31 | 3 | 1 | 35 |
| *Fast translation* | 18 | 11 | 29 | 3 | 1 | 33 |
| *Rotation* | 19 | 13 | 32 | 3 | 1 | 36 |
| *Fast rotation* | 19 | 13 | 32 | 3 | 1 | 36 |
| *Facial expressions* | 18 | 13 | 31 | 3 | 1 | 35 |
| *Total average* | 18 | 13 | 31 | 3 | 1 | 35 |

As you can see, the total average computation time to compute the head pose and capture the mouth and eyebrow movements is 35 ms. This means that the tracker will run 29 frames/second on average. The most expensive component is the optical flow pre-estimation with an average computation time of 18 ms. The main reason why the optical flow pre-estimation takes longer to compute than the template tracker is because not only does it has to estimate the pose, but it also has to track the points using the LK algorithm [41]. Tracking the two eyebrows and the mouth is really fast and together their only take 4 ms to compute.

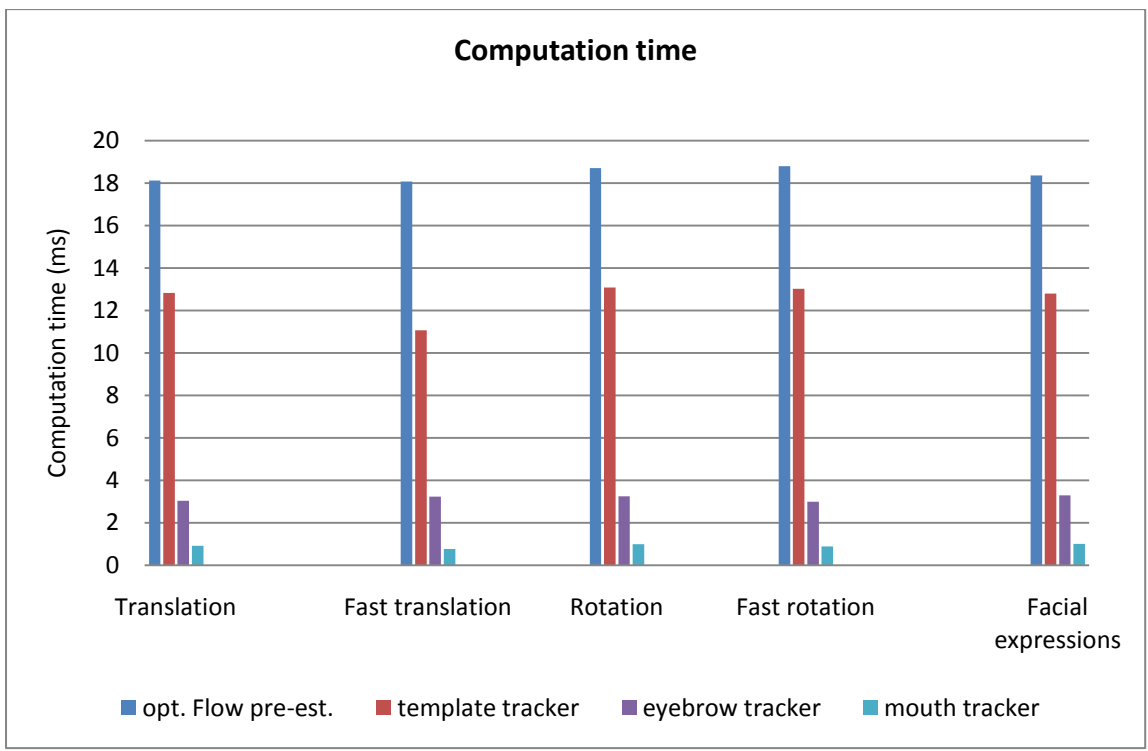

Figure 21: The average computation time (ms) of the different components.

From these results we can conclude that out total framework i.e. tracking the head, eyebrows and mouth, works in real-time. With an average of 29 frames/second it gives fluid tracking results. The accuracy of some of the components can even be improved by increasing the number of template points or increasing the rectified image size for the mouth and eyebrow tracker, and still keeping real-time performance.

# 6   Discussion & Future Work

We have presented a framework which allows us to track head movements and facial expressions. The head tracking experiment results show that our tracker can successfully track a head and its facial expressions. Experiments with the Boston University data set indicate that our tracker is capable of tracking the rotational movements of different subjects, but with a slightly higher error than the tracker of *Choi & Kim* [52]. But the results obtained with our own dataset gave significantly better results. This is due to the way the ground truth data is acquired and the type of perspective projection that is used. The Boston University ground truth data is acquired with a magnetic tracker, which gives more independent data than our semi-automatic annotation method. Also, we use a different perspective projecting which introduces some additional errors. Despite of these differences, when visually inspecting both experiments, the tracker is capable of tracking the head in a believable way i.e. when projecting the model onto the video, no significant errors are visible. Additional experiments show that out tracker is very good at handling fast horizontal, vertical and scaling movements. The combination of the optical flow pre-estimation and the template tracker allowed the tracker to handle fast moving subjects. It also reduced the number of head losses significantly when the subject is moving quickly. These improvements are most noticeable when having fast horizontal or vertical movements. When having fast rotations, the improvements are less noticeable. This understandable, because the pixel displacements are much less with fast rotations than with fast translations.

We also evaluated how robust our tracking is when the subject gets partially occluded and how much influence the outlier removal has on the robustness of our tracker. We found out that our tracker can keep track of the head, but only when a small part is occluded or when it is only occluded for a very short moment of time. We noticed that the template tracker is actually quite robust against occlusion, but unfortunately the combination with the optical flow pre-estimation makes the tracker more sensitive to occlusions. Most of the problems occur when the occluding object is moving. If this happens, it is possible that the optical flow points move along with the occluding object which will introduce errors in the pre-estimation. If the object is moving slow, it is not a problem because the optical flow points will be re-projected and updated. If the object is moving really fast, the outlier detection will detect the 'move along' points and won't include them in the pre-estimation. Unfortunately, including the outlier detection won't allow the head to be more occluded, but it does reduce the influence of the occlusion on the error estimate. It will reduce the number of head loss and therefore the number of re-initializations of the tracker.

The results from the mouth and eyebrow tracking experiment show that we are able to capture and track facial expressions in a way that is believable and recognizable. The eyebrow tracker gives a very robust tracking result, but it is possible that the tracker gets stuck in a local optimum, but when this happens it almost always recovers itself quickly. The experimental results show that also the mouth can be successfully tracked. One thing that we noticed is that the mouth tracker can sometimes be somewhat instable. It can sometimes 'vibrate' a little. This is probably due to the rectified image size. Increasing this size would probably reduce this 'vibration' effect. It is also sometimes possible that the tracker gives a wrong estimate due to rotation or occlusion. But overall, it does perform good enough to animate a 3D model in a way that looks natural.

With an average computation time of 35 ms to calculate the position and orientation of the head, mouth and eyebrows, our framework is very suitable for real-time applications. It is also possible to reduce the computation time even more if the code gets further optimized.

All the experiments are done using at least two completely different subjects and when we look at all the results, we see there is no real difference between the results of the different subjects. With these results we can say that our framework works with different subjects.

Future work may investigate how to improve the stability of the trackers. In most of the trackers, it may be beneficial to incorporate some kind of prediction model like the Kalman-filter. This should make the head tracker and the facial feature trackers more robust against occlusions an illumination changes. Also the optical flow pre-estimation can likely be improved by checking if the tracker points are really located on the head. This can be done with for example, simple template matching. An aspect that could be changed is the projection model. A full projection model will give more accurate results. To increase the robustness of the mouth tracker, a way to deal with a partial occluded mouth should be investigated. Also, the influence of the different parameters on the performance should be investigated.

# 7 Conclusion

In this thesis we proposed a complete framework for tracking a head and its facial movements. For the head tracker, we proposed an unique combination of two different approaches to overcome some of the difficulties in head tracking. We combined an optical flow based tracker with a template based tracker and exploit the strengths of both. We have also proposed two new and efficient approaches to track the mouth and eyebrow movements. This allows us to track and capture facial expressions.

The combination of an optical flow based pre-estimation for the template tracker showed to be very robust and effective when dealing with a fast moving subjects. The results show that using the pre-estimation reduces the tracking error and the number of head losses significantly. Because optical flow based approaches are known to be very prone to drifting we introduced a method to prevent this. When we know that the template tracker has found the head, we re-project and update the optical flow tracked points. This way, if there is some error accumulation, it will be corrected. With this method, there is no sign of drifting. To further improve the robustness of the optical flow tracker, we detect and remove erroneous tracked point from the tracking process. This makes the whole tracker more stable and also improves the performance of the tracker when dealing with fast movements. Another problem we had to deal with are partial occlusions of the head. For this, we introduced a way to detect and remove occluded template pixels. The results showed that our method did not directly allow more occlusions, but it made our tracked be less influenced by those occlusions.

An important part of our framework is the ability to track facial expression. To capture the facial expressions, we use two different trackers; one for the mouth and one for the eyebrows. To increase the efficiency and reduce the complexity of the tracking process, we use the acquired head pose estimate to create a rectified image of the mouth and eyebrow regions. Within the rectified images, we search for the mouth and eyebrows. Our mouth tracking method tries to reconstruct the initial mouth template instead of actually tracking the parts of the mouth. This way, we eliminated most of the mouths shape constrains and prevent mouth parts from drifting. The eyebrow tracker uses a different method. We track three eyebrow parts; the two eyebrow corners and the middle of the eyebrow. The tracking is done by minimizing an error function which does not only take template similarity information into account but also information about the shape of the eyebrow. Our results have shown that the mouth and eyebrow trackers can track the facial expression in a believable way. Unfortunately, the mouth tracker is not yet robust against occlusions. This also mean that large rotations can be a problem. But fortunately, the tracker is capable to correct itself after erroneous tracking.

One of the requirements of the framework is that it should be able to run in real-time. With an average of 35 ms per frame to compute the head pose and track the mouth and both the eyebrow i.e. 29 frames per second, we can state that it runs in real–time. Also, an important benefit of our framework is that changing the resolution of the input image does not affect the computation time of the trackers. This makes our framework very scalable.

Finally, we can conclude that our proposed methods works well. The combination of our methods results in a complete head and facial feature tracking framework that can handle different people with different faces and which can be used for real-time head tracking in all sorts of applications.